%% file: main.tex
\documentclass[10pt,twocolumn,letterpaper]{article}

\usepackage[pagenumbers]{cvpr}
\usepackage{amsmath,amsfonts}
\usepackage{amssymb}
\usepackage{algorithm}
\usepackage{algpseudocode}
\usepackage{array}
\usepackage{textcomp}
\usepackage{stfloats}
\usepackage{verbatim}
\usepackage{graphicx}
\usepackage{cite}
\usepackage{subcaption}
\usepackage{float}
\usepackage{multirow} 
\usepackage{booktabs} 
\usepackage{colortbl}
\definecolor{mygray}{gray}{.9}
\usepackage{amssymb}
\usepackage{pifont}
\newcommand{\lmm}[1]{\textcolor{black}{#1}}
\newcommand{\mm}[1]{\textcolor{black}{#1}}
\input{preamble}

\definecolor{cvprblue}{rgb}{0.21,0.49,0.74}
\usepackage[pagebackref,breaklinks,colorlinks,citecolor=cvprblue]{hyperref}

\def\paperID{*****} 
\def\confName{CVPR}
\def\confYear{2024}

\title{Qffusion: Controllable Portrait Video Editing via Quadrant-Grid \\Attention Learning}

\author{  \\[-25pt]
Maomao Li${^{*}}$,\quad Lijian Lin${^{*}}$,\quad Yunfei Liu${^{\dagger}}$, \quad Ye Zhu, \quad Yu Li${^{\dagger}}$ \\
International Digital Economy Academy (IDEA) \\
}

\begin{document}
\input{fig_tex/teaser}

\input{tex/abstract}
\input{tex/introduction}

\input{tex/related}
\input{tex/preliminary}
\input{tex/method}

\input{tex/experiment}

\input{tex/conclusion}

\small
\bibliographystyle{ieeenat_fullname}
\bibliography{main}


\clearpage
\input{sec/X_suppl}

\end{document}

%% file: preamble.tex
\usepackage[dvipsnames]{xcolor}


%% file: fig_tex/teaser.tex
\twocolumn[{%
\renewcommand\twocolumn[1][]{#1}%
\maketitle
\begin{center}
    \centering
    \captionsetup{type=figure}
\includegraphics[width=0.9\textwidth]{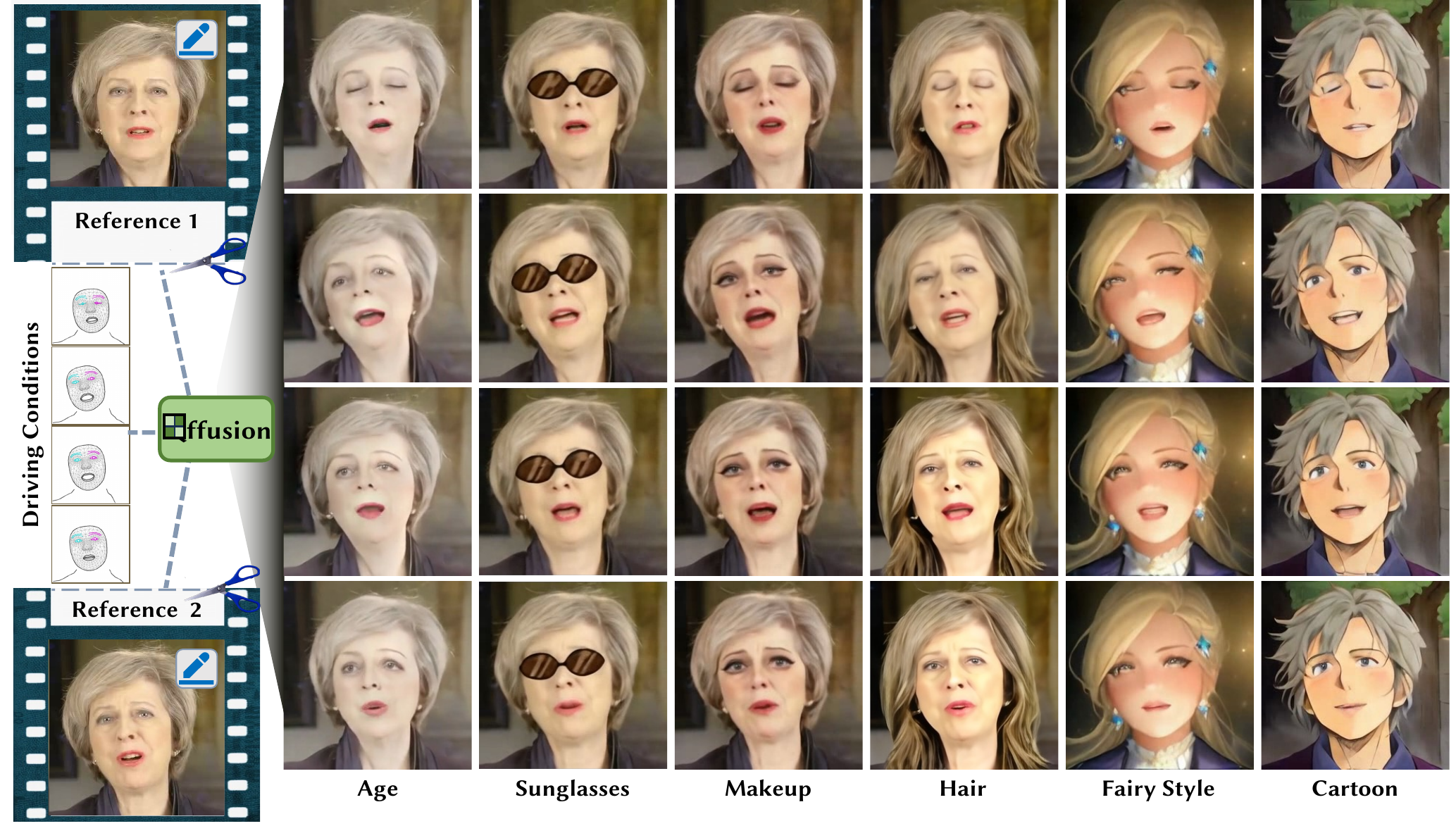}
\vspace{-2mm}
\caption{We present Qffusion, a simple yet effective \emph{dual-frame-guided} portrait video editing framework. Specifically, our Qffusion is trained as a general animation framework from two still reference images whereas it can perform portrait video editing effortlessly when using modified \emph{start} and \emph{end} video frames as references during inference. That is, we specify editing requirements by modifying two video frames rather than text. In this way, our Qffusion can perform fine-grained local editing (\eg, modifying age, makeup, hair, style, and wearing sunglasses).}
    \label{fig:teaser}
\end{center}%
}]

\let\thefootnote\relax\footnotetext{$*$ Eual contribution, 
$\dagger$ Corresponding authors.}

%% file: tex/abstract.tex
\begin{abstract}
This paper presents Qffusion, a dual-frame-guided framework for portrait video editing. Specifically, we consider a design principle of \textit{``animation for editing''}, and train Qffusion as a general animation framework from two still reference images while we can use it for portrait video editing easily by applying modified start and end frames as references during inference. Leveraging the powerful generative power of Stable Diffusion, we propose a Quadrant-grid Arrangement (\texttt{QGA}) scheme for latent re-arrangement, which arranges the latent codes of two reference images and that of four facial conditions into a four-grid fashion, separately. Then, we fuse features of these two modalities and use self-attention for both appearance and temporal learning, where representations at different times are jointly modeled under \texttt{QGA}. Our Qffusion can achieve stable video editing without additional networks or complex training stages, where only the input format of Stable Diffusion is modified. Further, we propose a Quadrant-grid Propagation (\texttt{QGP}) inference strategy, which enjoys a unique advantage on stable arbitrary-length video generation by processing reference and condition frames recursively. Through extensive experiments, Qffusion consistently outperforms state-of-the-art techniques on portrait video editing. \mm{Project page: \url{https://qffusion.github.io/page/}}.
\end{abstract}

%% file: tex/introduction.tex
\section{Introduction}
\label{sec:intro}
With the rapid proliferation of mobile internet and video platforms, portrait video editing has become one of the cornerstones of computer graphics and vision. Traditional approaches often require professional designers with time-consuming processes, \eg, scene setup, staged recording, editing, and repetitive iterations, which are laborious and inefficient. Considering fruitful endeavors have been pursued in image generation~\cite{dalle2,imagen,ldm,controlnet} and video generation~\cite{svd,pika,sora,gen-3} in recent years, employing powerful large-scale generative models to assist portrait video editing has become feasible.

Generally, the techniques for portrait video editing can be classified into two categories. One category is based on Generative Adversarial Networks (GANs)~\cite{goodfellow2014generative,pang2021image}. While these methods provide decent processing speeds, they often fail to generalize to unseen humans and suffer from unstable training~\cite {wu2020improving}. The other category comprises diffusion-based methods~\cite{ddpm,ddim}. These models~\cite{tune-a-video,pix2video,fatezero,tokenflow,li2023video, anyv2v} are pretrained on large image and video datasets, making them easily model arbitrary objects. Despite the generalizability and controllability of these diffusion models, they still face several challenges, which are listed in Tab.~\ref{tab:setting}. (i) The prevailing text-driven video editing methods~\cite{tune-a-video,pix2video,fatezero,tokenflow,li2023video} struggle to deal with specific local manipulations (\eg, hair editing) since text-driven editing cannot capture sufficient editing details directly. (ii) The very recent method AnyV2V~\cite{anyv2v} performs first-frame-guided editing, which first employs an off-the-shelf image editing model~\cite{brooks2023instructpix2pix,Zhang2023MagicBrush} to modify the first frame and then utilizes an image-to-video (I2V) generation model ~\cite{zhang2023i2vgen} to propagate such modifications. However, it often leads to a degraded quality, as a single edited frame cannot enforce sufficient appearance coherence. (iii) The video length of the existing methods is always constrained by limited computational resources.


To handle the above challenges, this paper proposes a \emph{dual-frame-guided} portrait video editing method dubbed Qffusion, which allows for fine-grained local editing on arbitrary-long videos. 
\lmm{Given that the existing first-frame-guided editing method~\cite{anyv2v} faces degraded generation quality, we argue that bi-directional context, rather than strictly causal dependencies, is fundamental for maintaining video quality.}
Specifically, we consider an ``animation for editing'' design principle and train Qffusion as a video animation network from two still reference images, which can perform portrait video editing effortlessly when applying edited start and end video frames as references during inference. That is, our Qffusion specifies editing requirements by modifying two video frames rather than text or one single frame. For example, we can use professional software (e.g., Photoshop or Meitu$^1$\footnote{$^1$\href{https://www.meitu.com/}{https://www.meitu.com/}}) for controllable and consistent reference frame editing.

Concretely, we design a {Quadrant-grid Arrangement} (\texttt{QGA}) scheme into image models (\ie, Stable Diffusion~\cite{ldm}) for video modeling, which only modifies the input format for training. Here, a four-grid representation is designed for two reference images and four sequential driving keypoints, respectively. In detail, we organize two reference images and two all-zero placeholders as intermediate masks into a big four-grid image. Then, this four-grid image representation is stacked with the corresponding four-grid driving representation. 
Benefiting from the feature aggregation ability of the attention mechanism, \texttt{QGA} scheme can establish the correspondence between driving conditions and reference appearance, where temporal clues are also modeled as motion information is embraced naturally in the four-grid driving representation. Moreover, to make an even motion modeling during inference, we design a Quadrant-grid Propagation (\texttt{QGP}) inference algorithm, which recursively uses generated frames at the current inference iteration as reference frames for the next iteration, making the edited video length unconstrained.

As shown in Fig.~\ref{fig:teaser}, Qffusion delivers impressive results in fine-grained local editing, \eg, adding sunglasses, editing age, hair, and style. Besides, since Qffusion is trained as a general video animation framework, we can flexibly use it for other applications, such as whole-body driving~\cite{animate} and jump cut smooth~\cite{jumpcut}. Our main contributions are:
\begin{itemize}
    \item We propose a novel dual-frame-guided framework for portrait video editing, which propagates fine-grained local modification from the start and end video frames.

    \item We propose a Quadrant-grid Arrangement (\texttt{QGA}) scheme to re-arrange reference images and driving signals under a four-grid fashion separately, which models appearance correspondence and temporal clues all at once. 

    \item We propose a recursive inference strategy named Quadrant-grid Propagation (\texttt{QGP}), which can stably generate arbitrary-long videos.

    \item Our Qffusion can deliver rich application extensions, \eg,  portrait video editing, whole-body driving~\cite{animate},  and jump cut smooth~\cite{jumpcut}, showing more competitive results with those state-of-the-art task-specific methods.

\end{itemize}

\input{table_tex/setting}

%% file: table_tex/setting.tex

\begin{table}[t]
\setlength\tabcolsep{1pt}
    \centering
    \small
    \begin{tabular}{l|c|c|c}
    \toprule  
      Settings   &\emph{Fine-grained} &\emph{Arbitrary-long} &\emph{Non-degraded appearance} \\
    \hline
 \emph{Text} & \ding{55}  &\ding{55}  &\ding{51}\\
\hline
\emph{First-frame}  &\ding{51}  &\ding{55} &\ding{55} \\
\hline
 \rowcolor{mygray} \emph{Dual-frame}
 &\ding{51}  &\ding{51}  &\ding{51}\\
     \bottomrule
    \end{tabular}    
    \caption{Categories and characteristics of diffusion-based video editing methods. This paper is the first to propose dual-frame settings for high-quality portrait video editing.}
\label{tab:setting}
\end{table}

%% file: tex/related.tex
\section{Related Works}
\noindent{\textbf{Diffusion Model for Image Generation and Editing.}}
Recently, diffusion models~\cite{ddpm,ddim} have emerged as a popular paradigm for text-to-image (T2I). DALLE-2~\cite{dalle2} and Imagen~\cite{imagen} can generate high-resolution images via cascaded diffusion models. Then, Stable Diffusion~\cite{ldm} proposes to train diffusion models in the learned latent space for less computational complexity. 
As for image editing, the early techniques~\cite{gafni2022make,nichol2021glide,dalle2,avrahami2023spatext,mokady2022self} need an editing mask provided by the user, which is time-consuming. 
To deal with this, there is a line of research conducts text-only image editing~\cite{p2p,pnp,prompt,null,cao2023masactrl,2022text2live,sdedit}, which changes the visual content of the input image following the target prompt without masks. Moreover, InstructPix2Pix~\cite{brooks2023instructpix2pix} and MagicBrush~\cite{Zhang2023MagicBrush} perform editing following human instructions. 
In addition, a group of methods~\cite{custom-diffusion,dreamartist,fastcomposer,dreambooth,textual-inversion,celeb} conduct personalized text-guided image editing and synthesize novel renditions of several given subjects in different contexts. 
To further improve spatial controllability, ControlNet~\cite{controlnet} introduces a side path to Stable Diffusion to accept extra conditions like edges, depth, and human pose.

\noindent{\textbf{Diffusion Model for Video Generation and Editing.}}
Following image generation and editing diffusion models, there have also been substantial efforts in Text-to-Video (T2V). Besides operating diffusion process directly on pixel space~\cite{imagen-video,singer2022make}, the recent T2V models~\cite{blattmann2023align,zhou2022magicvideo,esser2023structure,wang2023modelscope} draw inspiration from Stable Diffusion~\cite{ldm} and generate high-quality videos via a learned latent space.
Apart from text-driven video generation, several representative works~\cite {svd,pika,sora,gen-3,zhang2023i2vgen} lay the cornerstone for image-to-video (I2V) generation.

Regarding text-driven video editing, Tune-A-Video~\cite{tune-a-video} first proposes an efficient one-shot tuning strategy based on Stable Diffusion.
Then, a group of methods~\cite{tokenflow,li2023video,pix2video,fatezero,cong2023flatten,yatim2023space} conduct zero-shot video editing, where various attention mechanisms are designed to capture temporal cues without extra training. 
Further, following the substantial efforts of I2V generation, AnyV2V~\cite{anyv2v} conducts image-driven video editing by propagating modified content from the first edited frame. 
However, it leads to a degraded quality, where those frames that are far away from the first frame usually present an unpleasing reconstruction and editing. \lmm{The very recent method Go-with-the-flow~\cite{burgert2025go} also allows for first-frame-guided editing, which replaces random temporal Gaussianity with correlated warped noise derived from optical flow fields. Although promising performance, it still can only edit limited-frame videos, since the relied I2V model is trained on a limited number of frames. In contrast, our Qffusion can perform arbitrary-long portrait video editing.}

Generally, text-driven video editing methods struggle with certain fine-grained local manipulations on portrait videos, \eg, modifying hair. Different from the typical text-driven video editing methods, Codef~\cite{codef} introduces a new type of video representation, which consists of a canonical content field and a temporal deformation field recording static contents and transformations separately. By editing the canonical image, Codef can carry out fine-grained local editing. However, it needs training on each video to be edited, whereas ours is a general framework once it is finished training.

\noindent{\textbf{Diffusion-based Video Animation.}}
In recent years, apart from pose-controllable text-to-video generation~\cite{ma2024follow}, some researchers focus more on generating animated videos from \textit{still images} with diffusion models. 
DreamPose~\cite{dreampose} proposes a two-stage finetuning strategy with pose sequence. BDMM~\cite{bdmm} designs a Deformable Motion Modulation that utilizes geometric kernel offset with adaptive weight modulation for subtle appearance transfer. Besides, Animate Anyone~\cite{animate} temporally maintains consistency by a ReferenceNet merging detail features via spatial attention, where a pose-guided module is designed for movements.
Unlike them, our Qffusion is very flexible for various applications, such as portrait video editing, whole-body driving~\cite{animate}, and jump cut smooth~\cite{jumpcut}.

%% file: tex/preliminary.tex
\section{Preliminary of Stable Diffusion}

As a powerful image synthesis model, Stable Diffusion~\cite{ldm} consists of a VAE~\cite{vae}, a diffusion process, and a denoising process. Here, VAE provides a learnable latent space, avoiding the massive resources required for pixel-level calculation.

\noindent \textbf{Diffusion Process.} In the diffusion process, the model progressively corrupts input data  $\mathbf{z}_0 \sim p(\mathbf{z}_0)$ according to a predefined schedule $\beta_t \in (0, 1)$, turning data distribution into an isotropic Gaussian in $T$ steps. Formally, it can be expressed as:
\begin{align}
q\left(\mathbf{z}_{1: T} \mid \mathbf{z}_{0}\right) & = \prod q\left(\mathbf{z}_{t} \mid \mathbf{z}_{t-1}\right), \qquad {t \in [1,..., T]}.
\end{align}
\noindent \textbf{Denoising Process.} 
In the denoising process, the model learns to invert the diffusion procedure so that it can turn noise into real data distribution at inference.
The corresponding backward process can be described as follows:
\begin{equation} 
\begin{aligned}
\label{eq:denoising}
p_{\theta}(\mathbf{z}_{t-1}| \mathbf{z}_{t}) &= \mathcal{N}(\mu_{\theta}(\mathbf{z}_{t}, t), \Sigma_{\theta}(\mathbf{z}_{t}, t)) \\&= \mathcal{N}(\frac{1}{\sqrt{\alpha_{t}}}(\mathbf{z}_{t}-\frac{\beta_{t}}{\sqrt{1-\bar{\alpha}_{t}}} {\epsilon}),
\frac{1-\bar{\alpha}_{t-1}}{1-\bar{\alpha}_{t}} \beta_{t}),
\end{aligned}
\end{equation}
where ${\epsilon} \sim \mathcal{N}(\mathbf{0},\mathbf{I})$, $\alpha_{t}=1-\beta_{t}$, $\bar{\alpha}_{t}=\prod_{i=1}^{t} \alpha_{i}$ and $\theta$ denotes parameters of the denoising neural network. The training objective is to maximize the likelihood of observed data $p_{\theta}\left(\mathbf{z}_{0}\right)=\int p_{\theta}\left(\mathbf{z}_{0: T}\right) d \mathbf{z}_{1: T}$, by maximizing its evidence lower bound (ELBO), which effectively matches the true denoising model $q\left(\mathbf{z}_{t-1} \mid \mathbf{z}_{t}\right)$ with the parameterized $p_{\theta}\left(\mathbf{z}_{t-1} \mid \mathbf{z}_{t}\right)$. During training, the denoising network $\mathbf{\epsilon}_\theta(\cdot)$ restore $\mathbf{z}_0$ given any noised input $\mathbf{z}_t$, by predicting the added noise $\epsilon$ via minimizing the noise prediction error: 
\begin{align}
\mathcal{L}_{t} & \!=\! \mathbb{E}_{\mathbf{z}_{0}, \epsilon \sim \mathcal{N}(\mathbf{0},\mathbf{I}) }\left[\left\|{\epsilon}\!-\!\mathbf{\epsilon}_{\theta}\left(\sqrt{\bar{\alpha}_{t}}\mathbf{z}_0\!+\!\sqrt{1-\bar{\alpha}_{t}} {\epsilon}; t\right)\right\|^{2}\right]. \label{equ:noise_loss}
\end{align}

To make the model conditioned on extra condition $\mathbf{z}_c$, we can inject $\mathbf{c}$ into $\mathbf{\epsilon}_\theta(\cdot)$ by replacing 
$\mu_{\theta}\left(\mathbf{z}_{t}, t\right)$ and $\Sigma_{\theta}\left(\mathbf{z}_{t}, t\right)$ with $\mu_{\theta}\left(\mathbf{z}_{t}, t, \mathbf{c}\right)$ and $\Sigma_{\theta}\left(\mathbf{z}_{t}, t,\mathbf{c}\right)$. 

%% file: tex/method.tex
\input{fig_tex/overall}
\section{Methods}
\label{sec:method}
This paper proposes a \emph{dual-frame-guided} portrait video editing method dubbed Qffusion, which can perform fine-grained or local editing on arbitrary-long videos. Specifically, we consider an ``animation for editing'' principle, and train Qffusion as a video animation framework from two still reference images while we can use it for portrait video editing easily by applying edited start and end video frames as references during inference. That is, we first specify editing requirements by modifying the start and end frames with professional software and then use the proposed Qffusion to propagate these fine-grained local modifications to the entire video.

The Section is organized as follows: Sec.~\ref{4.1} first introduces an overview of our proposed Qffusion. Sec.~\ref{4.3} illustrates the Quadrant-grid Arrangement (\texttt{QGA}) scheme in SD for latent re-arrangement. Then, our recursive inference strategy Quadrant-grid Propagation (\texttt{QGP}) for stable and arbitrary-length video generation is presented in Sec.~\ref{4.4}.

\subsection{Overview}
\label{4.1}
The pipeline of Qffusion is illustrated in Fig.~\ref{fig:pipeline}. Like SD, our model consists of two parts, i.e., VAE and latent diffusion. In Qffusion, we propose a \texttt{QGA} scheme to arrange four sequential frames into a large four-grid image, where the upper-right and bottom-left frames are masked for generation. This four-grid images are then stacked with their corresponding four-grid keypoints, forming a composite input that encodes both visual and motion information. Formally, given four sequential frames $\{\mathbf{I}^a, \mathbf{I}^b, \mathbf{I}^c, \mathbf{I}^d\}$ and their corresponding condition image (\ie, keypoints) $\{\mathbf{C}^a, \mathbf{C}^b, \mathbf{C}^c, \mathbf{C}^d\}$, we replace $\mathbf{I}^b, \mathbf{I}^c$ with all-zero masks and train our model to reconstruct them. The VAE encoder $\mathcal{E}$ first encodes $\mathbf{I}^a, \mathbf{I}^d, \mathbf{C}^a, \mathbf{C}^b, \mathbf{C}^c, \mathbf{C}^d$ into latent codes $\mathbf{r}^a, \mathbf{r}^d, \mathbf{c}^a, \mathbf{c}^b, \mathbf{c}^c, \mathbf{c}^c$, respectively. 
Then, these input latent codes are combined with a noise map to form a fused code through our \texttt{QGA} scheme. Next, a denoiser $\mathbf{\epsilon}_{\theta}$ learns driving correspondence and temporal clues from these latent codes and predicts the denoised latent. 
Finally, a VAE decoder $\mathcal{D}$ decodes the denoised latent into images corresponding to the input conditions.
The process of Qffusion is: 
\begin{equation}\label{eq:qffussion}
    \tilde{\mathbf{I}}^b, \tilde{\mathbf{I}}^c = \text{Qffusion}(\mathbf{I}^a, \mathbf{I}^d, \mathbf{C}^a, \mathbf{C}^b, \mathbf{C}^c, \mathbf{C}^d).
\end{equation}
In summary, Qffusion takes two frames $\mathbf{I}^a$ and an $\mathbf{I}^d$ as appearance references, where the condition images of references $\mathbf{C}^a$ and $\mathbf{C}^d$ and that of intermediate frames ($\mathbf{C}^b$ and $\mathbf{C}^c$) as motion signals for the generation of $\tilde{\mathbf{I}}^b$ and $\tilde{\mathbf{I}}^c$. 
After training, the model would generate two portrait frames at each inference time. 
By replacing the intermediate conditions sequentially, our method can generate arbitrary-length videos easily.

\subsection{Quadrant-grid Arrangement}
\label{4.3}
Based on SD, we train our Qffusion as a general video animation framework from two reference images and four driving signals. 
Specifically, we propose a Quadrant-grid Arrangement (\texttt{QGA}) scheme to establish the correspondence between two modalities (i.e., appearance features and driving signals) for appearance consistency in the denoiser UNet $\mathbf{\epsilon}_{\theta}$, while 
jointly modeling the temporal clues at different times. 

\texttt{QGA} arranges the latent codes of reference images ($\mathbf{r}^a$ and  $\mathbf{r}^d$) and two all-zero placeholder masks for intermediate frames into a big four-grid image $\{\mathbf{r}^a, \mathbf{0}, \mathbf{0}, \mathbf{r}^d\}$, where reference images are assigned to upper-left and bottom-right locations. Similarly, we combine four driving conditions into a big four-grid condition image as $\{\mathbf{c}^a$, $\mathbf{c}^b$, $\mathbf{c}^c, \mathbf{c}^d\}$, which is stacked with the previous four-grid appearance latents $\{\mathbf{r}^a, \mathbf{0}, \mathbf{0}, \mathbf{r}^d\}$. Here, a one-to-one correspondence between the appearance and conditions is achieved. In this way, appearance representations of different frames would establish spatial relationships in the self-attention layers for the reconstruction of $\tilde{\mathbf{I}}^b$ and $\tilde{\mathbf{I}}^c$. In addition, temporal clues are also modeled naturally in \texttt{QGA} since motion information is embraced naturally in the composed four-grid representation.

As illustrated in Fig.~\ref{fig:pipeline}, \texttt{QGA} scheme stacks the four-grid representations of reference frames ($\mathcal{R}$) and that of condition images ($\mathcal{C}$), and then combines a same-size noise map ($\mathcal{Z}_t$), thus obtaining the fused latent representation $\mathcal{Q}_t$ as:
\begin{equation} 
\begin{aligned}
\label{eq:qga}
    \mathcal{Q}_t &= \texttt{QGA}(\mathbf{I}^a,\mathbf{C}^a, \mathbf{C}^b, \mathbf{C}^c, \mathbf{C}^d, \mathbf{I}^d, t) 
                 = \mathcal{R} \odot \mathcal{C} \odot \mathcal{Z}_t \\
                 &= \left[\begin{array}{cc}
                     \mathbf{r}^a & \mathbf{0} \\
                     \mathbf{0} & \mathbf{r}^d
                 \end{array} 
                   \right] \odot
                 \left[\begin{array}{cc}
                     \mathbf{c}^a & \mathbf{c}^b \\
                     \mathbf{c}^c & \mathbf{c}^d
                 \end{array} 
                   \right] \odot
                 \left[\begin{array}{cc}
                     \mathbf{z}_t^a & \mathbf{z}_t^b \\
                     \mathbf{z}_t^c & \mathbf{z}_t^d
                 \end{array} 
                   \right],
\end{aligned}
\end{equation}
where $\mathbf{c}^b$ and $\mathbf{c}^c$ are driving latent codes from intermediate condition images, i.e., $\mathbf{c}^b=\mathcal{E}(\mathbf{C}^b)$ and $\mathbf{c}^c=\mathcal{E}(\mathbf{C}^c)$. $\mathbf{z}_t^*$ is the noise of $*$-th frame at timestep $t$. 
$\odot$ denotes channel-wise concatenation. Each latent code in the inputs of \texttt{QGA} (\ie, $\mathbf{r}^*, \mathbf{c}^*, \mathbf{z}^*$) is with the size of $\mathbb{R}^{B\times C\times H\times W}$ and thus $\mathcal{Q}_t \in \mathbb{R}^{B\times 3C\times 2H\times 2W}$, where $B$ denotes batch size. 


Next, the fused $\mathcal{Q}_t$ is fed to the denoiser $\mathbf{\epsilon}_{\theta}$ to predict the denoised latent codes. During training, the diffusion process iteratively adds noises to $\mathcal{Z}_0$ and eventually leads to $\mathcal{Z}_T$. In the denoising process, the denoiser $\mathbf{\epsilon}_{\theta}$ aims to recover latent codes $[[\mathbf{z}^a, \mathbf{z}^b]^\top, [\mathbf{z}^c, \mathbf{z}^d]^\top]$ based on the fused latent $\mathcal{Q}_T$.

Utilizing the learning framework in SD, our method only adjusts the number of I/O channels of UNet. After the denoising process, we can obtain the generated intermediate frame $\tilde{\mathbf{I}}^i = \mathcal{D}(\tilde{\mathbf{z}}^i)$, where $i \in \{b, c\}$ and $\tilde{\mathbf{z}}^i$ is generated by splitting and unstacking the denoised $\tilde{\mathcal{Q}}_0 = \epsilon_\theta(\tilde{\mathcal{Q}}_{1}), \dots, \tilde{\mathcal{Q}}_{T-1} = \epsilon_\theta(\mathcal{Q}_T)$:
\begin{equation} \label{eq:unstack}
    \tilde{\mathbf{z}}^i = \tilde{\mathcal{Z}}_0[:, : ,: H, W:], ~~s.t.~~
    \tilde{\mathcal{Z}}_0 = \tilde{\mathcal{Q}}_0[:, 2C:, :]. 
\end{equation}
In this way, our training objective can be expressed as:
\begin{equation}
    \mathcal{L}_{t}^\prime \!=\! \mathbb{E}_{\mathcal{Z}_{0}, \mathcal{R}, \mathcal{C}, \epsilon \sim \mathcal{N}(\mathbf{0},\mathbf{I}) }\!\!\left[\!\left\|{\epsilon}\!-\!\mathbf{\epsilon}_{\theta}\!\!\left(\sqrt{\bar{\alpha}_{t}}\mathcal{Z}_0\!+\!\sqrt{1-\bar{\alpha}_{t}} {\epsilon}; t, \mathcal{R}, \mathcal{C}\right)\!\right\|^{2}\!\right].
\nonumber
\end{equation}

\subsection{Quadrant-grid Propagation for Inference}
\label{4.4}





\input{fig_tex/inference}
The remaining problem is how to continuously generate all intermediate frames given the start and end reference frames and driving signals during inference. Considering our quadrant-grid training design, we also maintain this fashion for inference. Specifically, we only generate two frames for each inference time. Then, we can generate arbitrary-length videos easily by replacing the intermediate conditions sequentially. This makes portrait video animation and editing no longer constrained by limited computing resources.

\noindent{\textbf{Naive Inference.}}
We assume an input portrait video with $K+1$ frames, which are indexed are $[0,1,2,...K]$ respectively. A naive inference way is \textbf{fixing} two reference images as 0-th and $K$-th frames to gradually generate intermediate frames, ie, $\{1,K-1\}, \{2,K-2\},...,\{\left \lfloor{K/2} \right \rfloor, \left \lfloor {(K+1)/2} \right \rfloor \}$. This requires $K/2$ inference times. However, the synthesized frames suffer from the issue of excessive interval: (1) the interval between two generated frames is excessive, e.g., a $K-2$ gap between the synthesized 1-th and $(K-1)$-th frame, (2) the intervals between intermediate frames and reference frames are excessive, e.g., a $\left \lfloor {K/2} \right \rfloor$ gap between the synthesized $\left \lfloor{K/2} \right \rfloor$ frame and $0$-th reference frame. The issue would lead to unstable motion modeling, especially when $K$ is big, making the naive inference a suboptimal solution.

\noindent{\textbf{\texttt{QGP} Inference.}}
In order to seek a more even motion modeling, this paper proposes a recursive influence strategy Quadrant-grid Propagation (\texttt{QGP}), in which we use generated frames at the current inference iteration
as reference frames for the next iteration.
As shown in Fig.~\ref{fig:infer_pipe}, we first use 0-th and $K$-th reference frames to generate the intermediate two frames, which are indexed as $\left \lfloor \frac{1}{3}K\right \rfloor$-th and $\left \lfloor \frac{2}{3}K\right \rfloor$-th. 
Then in the next iteration, the newly generated 
$\left \lfloor \frac{1}{3}K\right \rfloor$-th and $\left \lfloor \frac{2}{3}K\right \rfloor$-th  
frames would serve as reference frames for their corresponding intermediate frames. 
The process is stopped until all $K+1$ frames are synthesized.
Specifically, the quadrant arrangement at each inference iteration would be: $\left \{\left[0, \left \lfloor \frac{K}{3}\right \rfloor, \left \lfloor \frac{2K}{3}\right \rfloor, K \right] \right \},  \{\left[0, \left \lfloor \frac{K}{9}\right \rfloor, \left \lfloor \frac{2K}{9}\right \rfloor, \left \lfloor \frac{K}{3}\right \rfloor \right]$, $\left[\left \lfloor\frac{K}{3}\right \rfloor, \left \lfloor\frac{4K}{9}\right \rfloor, \left \lfloor\frac{5K}{9}\right \rfloor, \left \lfloor\frac{2K}{3}\right \rfloor \right],
\left[\left \lfloor\frac{2K}{3}\right \rfloor, \left \lfloor\frac{7K}{9}\right \rfloor, \left \lfloor\frac{8K}{9}\right \rfloor,K \right] \, \}$, $..., \{[0, 1, 2, 3],...\}$. Note that at each iteration, \texttt{QGP} makes the intervals between each sub-image in the four-grid representation the same, leading to a more even temporal sampling than naive inference.

Summing up, although both the proposed \texttt{QGP} and the naive inference can synthesize arbitrary-length videos, the former can carry out smoother temporal modeling by making the interval of generated frames the same. More implementation details can be found in our Appendix.

%% file: fig_tex/overall.tex
    \begin{figure*}[t]
      \centering
\includegraphics[width=0.95\linewidth]{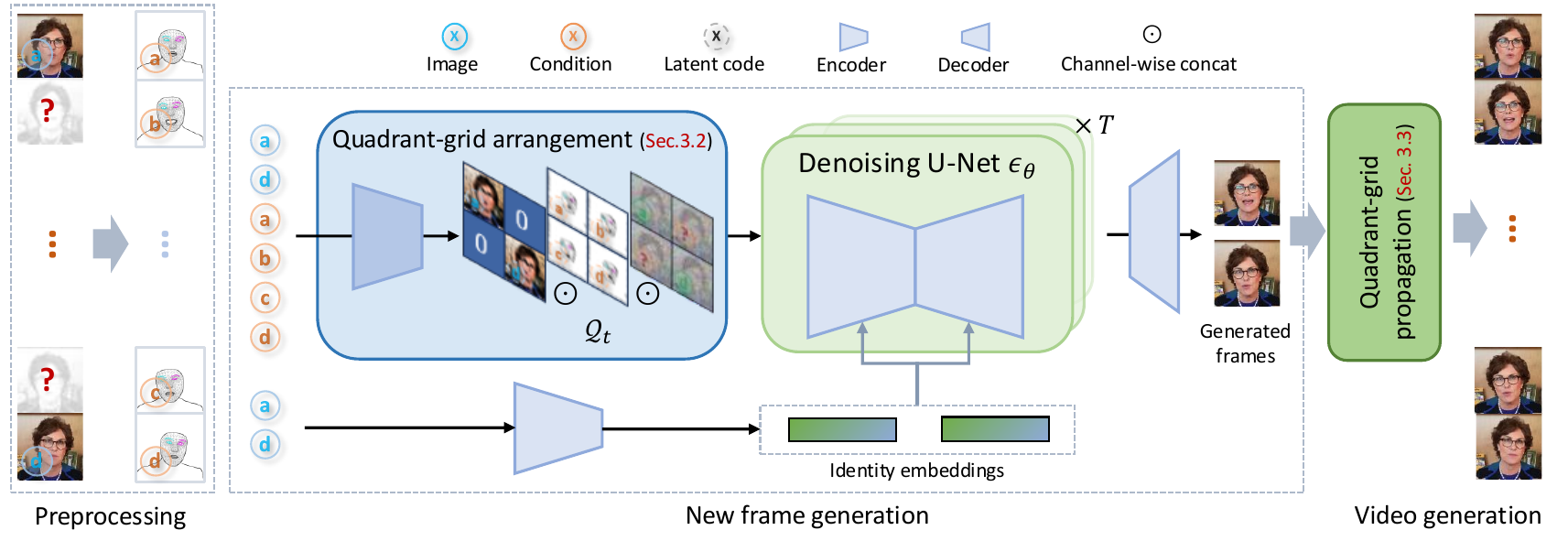}
       \caption{Overview illustration of Qffusion. As for training, we first design a Quadrant-grid Arrangement (\texttt{QGA}) scheme for latent re-arrangement, which arranges the latent codes of two reference images and that of four portrait landmarks into a four-grid fashion, separately.  Then, we fuse features of these two modalities and use self-attention for both appearance and temporal learning. Here, the facial identity features~\cite{deng2019arcface} are also put into cross-attention mechanism in the denoising U-Net for further identity constraint. During inference, a stable video is generated via our proposed Quadrant-grid Propagation (\texttt{QGP}) strategy.}
       \label{fig:pipeline}
    \end{figure*}

%% file: fig_tex/inference.tex
\begin{figure}
    \centering
\includegraphics[width=1.0\linewidth]{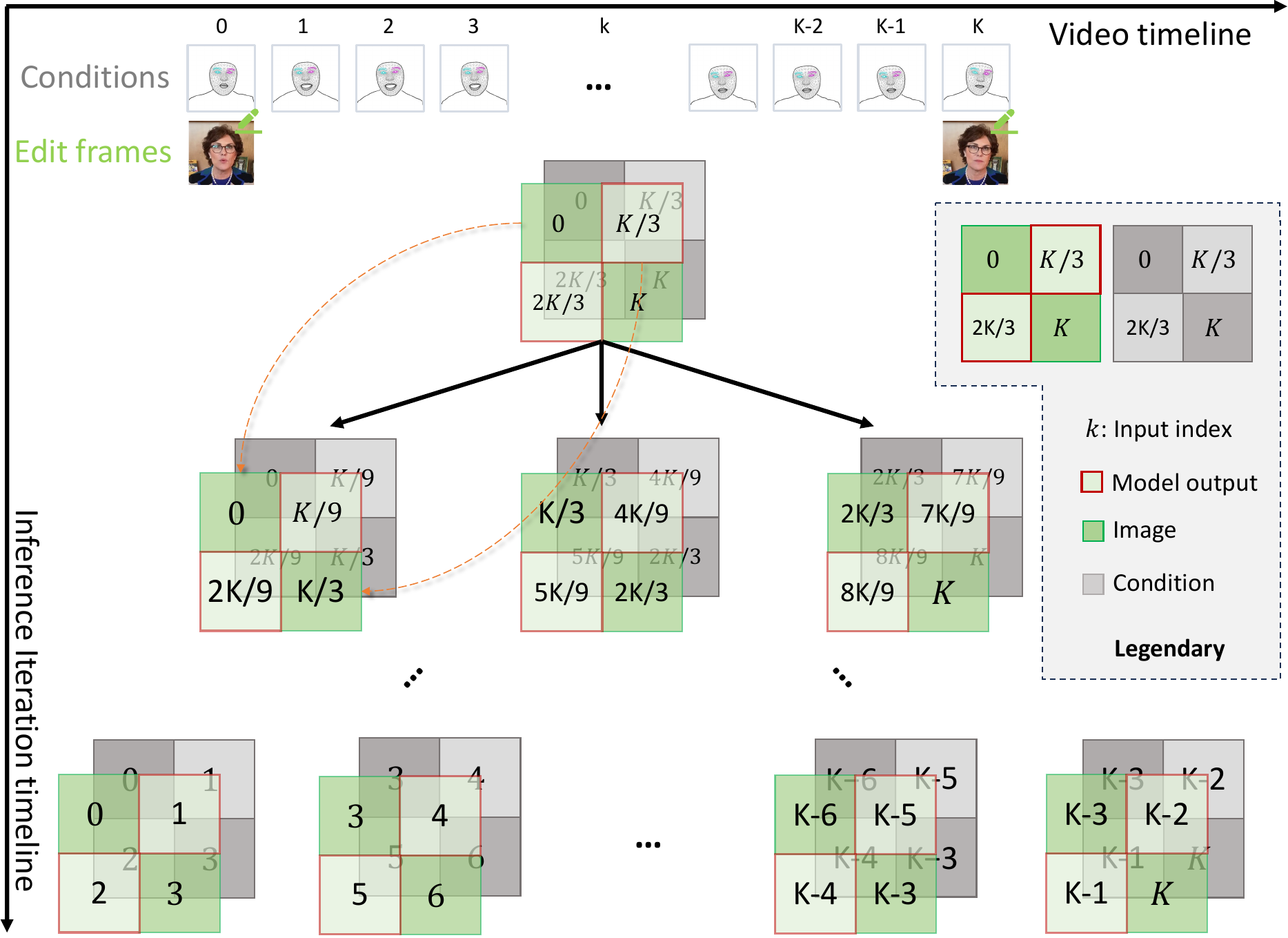}
    \caption{The illustration of Quadrant-grid Propagation (\texttt{QGP}) for stable arbitrary-length video generation. At each iteration, we use the same intervals between each sub-images in the four-grid representation to bring even temporal modeling. We omit the rounding symbols for clarity.
    }
    \label{fig:infer_pipe}
\end{figure}

%% file: tex/experiment.tex
\section{Experiment}
\label{sec:exp}
\subsection{Implementation Details} \label{5.1}
\noindent{\textbf{Training.}}
We use a fixed interval of 5 on video sequences to collect four frames as $\{\mathbf{I}^a, \mathbf{I}^b, \mathbf{I}^c, \mathbf{I}^d\}$ in the \texttt{QGA} scheme. Based on SD 1.5, Qffusion uses an AdamW optimizer~\cite{Ilya2017AdamW} with gradient accumulations set to 2. For the learning rate schedule, a warm-up strategy is applied, which gradually increases the learning rate to 0.0001 throughout 10,000 steps. Other hyper-parameters are followed with SD. 
We train all of our models on an NVIDIA A100 GPU, where a VAE and a denoiser $\epsilon_\theta$ are trained, respectively. For VAE training, the batch size is set to 4. The VAE training takes about 8 hours. For denoiser training, the batch size is set to 1. The denoiser training takes about 8 hours.



\input{table_tex/quantitative}

\input{table_tex/ablation}

\noindent{\textbf{Experimental setup.}}
We train Qffusion on HDTF dataset~\cite{zhang2021HDTF} and evaluate it on LSP~\cite{lu2021live} and some videos in RAVDESS~\cite{RAVDESS} and Celebv-HQ~\cite{zhu2022celebvhq} datasets. Each video contains a high-resolution portrait. The average video length is 1-5 minutes processed at 25 fps. Each video is cropped to keep the face at the center and resized to $256 \times 256$. LSP contains 4 video sequences. Our conditions consist of dense facial landmarks and torso lines. We detect 478 3D facial landmarks for all videos using Mediapipe~\cite{lugaresi2019mediapipe}. The 3D torso points describing the shoulder boundaries are estimated by~\cite{liu2023moda}.
Besides, we use professional software (e.g., Photoshop or Meitu) to edit start and end frames to maintain consistency.

\input{fig_tex/reconstruction}
\input{fig_tex/ablation}

\noindent{\textbf{Evaluation Metrics.}}
To verify the effectiveness of our Qffusion on portrait video animation, we use the average Peak signal-to-noise Noise Ratio (PSNR) \cite{PSNR}, Structural Similarity Index Measure (SSIM) \cite{SSIM}, and Learned perceptual similarity (LPIPS)~\cite{lpips}. Besides, we apply {Warp Error}~\cite{pix2video,tokenflow} to measure the temporal consistency of generated videos. Specifically, we first estimate optical flow~\cite{raft} of the input video and then use it to warp the generated frames. Next, the average MSE between each warped frame and the target ones is calculated.

To evaluate portrait video editing, we first use {CLIP-Image} similarity to measure the reference alignment of edited videos. It computes the average cosine similarity of image embeddings from CLIP model~\cite{radford2021learning} between the edited video frame and the rest of the generated frames. {Warp Error} is also leveraged to measure the temporal consistency of edited videos.


\subsection{Qualitative Comparison on Animation}
Since Qffusion is trained as a video animation framework, we first present a visual comparison with several competitive portrait video animation methods in Fig.~\ref{fig:qualitative}, which are: 
(i) {\textbf{GAN.}} A UNet-based GAN~\cite{lu2021live}, which is trained for 
reconstruction. Here, two reference images and condition images are used as input
to predict corresponding animation images.
(ii) \textbf{ControlNet.} 
We apply ControlNet~\cite{controlnet} to encode the condition images and the reference ones. Specifically,
we first use two "reference-only" ControlNet$^2$~\footnote{$^2$\href{https://github.com/Mikubill/sd-webui-controlnet/discussions/1236}{https://github.com/Mikubill/sd-webui-controlnet/discussions/1236}}
to encode two reference images, and an "OpenPose" ControlNet$^3$
~\footnote{$^3$\href{https://huggingface.co/lllyasviel/sd-controlnet-openpose}{https://huggingface.co/lllyasviel/sd-controlnet-openpose}} for condition encoding. Then, all encoded features are fed into SD to generate animated images that follow the condition motion and reference appearance. 
(iii) {{\textbf{ControlNet}} $+$ \textbf{AnimateDiff}.} We insert the temporal module of AnimateDiff~\cite{animatediff} into ControlNet for temporal consistency.
\mm{(iv) \textbf{Animate Anyone*.}} We use the re-produced version~\cite{animatemoore} of Animate Anyone~\cite{animate} for portrait video animation, where the official code is not publicly available. \mm{The re-produced \textbf{Animate Anyone} proposes a face reenactment method using the facial landmarks of driving video to control the pose of the given source image, and keeping the identity of the source image. This face reenactment model is trained on abundant portrait data.} (v) \textbf{MagicAnimate.} We employ MagicAnimate~\cite{xu2024magicanimate} for portrait video animation.

Although Qffusion can generate arbitrary-length videos, we use all methods to generate 80 frames for comparison here. \lmm{We additionally provide a 502-frame example in our Appendix for a long-term stability test.}
Fig.~\ref{fig:qualitative} shows the animation performance, where two reference images are omitted for simplicity. Our Qffusion has the most consistent appearance details and motion. Note that AnimateDiff only generates 16-32 frames. To generate 80 frames, we use an overlap generation strategy (i.e., overlapping 8 frames for a 16-frame generation) to maintain continuity.
\mm{In addition, \textbf{MagicAnimate} cannot perform portrait-video animation well. The reason is that the method is trained on whole-body data, which can not be generalized to cross-domain data such as portrait video.}


\input{fig_tex/inference_supp}

\subsection{Quantitative Comparison on Animation}
We report PSNR, SSIM, Warp Error, and LPIPS for quantitative comparison in Tab.~\ref{tab:quantitative}. Our Qffusion excels the current state-of-the-art methods by a large margin on PSNR, SSIM, and LPIPS. Although {\textbf{GAN}} can achieve a slightly superior Warp Error than our Qffusion, it yields the worst ID fidelity in Fig.~\ref{fig:qualitative}.
Besides, Qffusion achieves the best Warp Error among diffusion-based methods, which shows our capacity for long-term temporal modeling by Quadrant-grid Arrangement (\texttt{QGA}) and Quadrant-grid Propagation (\texttt{QGP}). Besides, although the temporal module of AnimateDiff can bring a temporal prior, it only supports fixed and limited length generation. Even if the overlap strategy is used for long video generation, the Warp Error of {{\textbf{ControlNet}} $+$ \textbf{AnimateDiff}} is still the worst.
\lmm{Further, the performance of \mm{\textbf{Animate Anyone*}} and \textbf{MagicAnimate} are also inferior to our method on portrait video animation.}

\subsection{Ablation Studies}  

\noindent{\textbf{Quadrant-grid Arrangement.}}
We conduct ablation studies to validate the effectiveness of our key design of \texttt{QGA}. For a fair comparison, we use the same training and inference strategies. 
We report the performance of the following three settings for latent arrangement. As illustrated in Tab.~\ref{tab:ablation}, (a) Two images side-by-side ($\{\mathbf{r}^a, \mathbf{0}\}$): a reference image is concatenated with an all-zero placeholder mask side by side. (b) Four-grid with one reference image ($\{\mathbf{r}^a, \mathbf{0}, \mathbf{0}, \mathbf{0}\}$): a reference image is arranged in the left-top corner of a four-square grid, leaving the remaining three squares to be zeros. (c) The proposed quadrant-grid design \texttt{QGA} ($\{\mathbf{r}^a, \mathbf{0}, \mathbf{0}, \mathbf{r}^d\}$).

Quantitative results of our quadrant-grid design are shown in Tab.~\ref{tab:ablation}.
Both (a) and (b) use only one reference image as a start point for generation. However, (a) and (b) yield inferior performance in temporal consistency and image quality for portrait video animation. Our \texttt{QGA}, on the contrary, uses two reference frames to constrain the generation of intermediate frames, which achieves significant gain over (a) and (b), demonstrating our effectiveness.

\input{table_tex/tab_inference}

We also perform a qualitative evaluation to verify the effectiveness of our quadrant-grid design in Fig.~\ref{fig:ablation1}.
The results of (a) exhibit noticeable lighting jitter ($1$-st row, $2$-nd column) and severe artifacts (in $1$-st row, $3$-rd and $4$-th column).
The results of (b) show color and lighting jitters among frames ($2$-nd row, $2$-nd column), and inaccurate mouth movement ($2$-nd row, $4$-th column).
As shown in the $3$-rd row, our \texttt{QGA} can generate temporal-consistent portrait videos.
To sum up, compared with one reference, we argue that two references can help regularize temporal appearance in the generated video.

\noindent{\textbf{Quadrant-grid Propagation.}} To validate the effectiveness of our \texttt{QGP} inference, we compare it with the naive inference in Tab.~\ref{tab:inference}.  \texttt{QGP} outperforms the naive inference on all metrics, 
especially on Warp Error, which decreases \textbf{72.6\%} dramatically. The excellent Warp Error of \texttt{QGP} shows its superiority in long-term temporal consistency, which achieves intervals between each sub-image more even in our quadrant-grid design. 
Besides, we provide the qualitative comparison between the proposed \texttt{QGP} inference and the naive inference in Fig.~\ref{fig:inference_supp}, which also supports our findings in Tab.~\ref{tab:inference}.

\input{fig_tex/video_editing}

\input{fig_tex/whole_body}
\input{table_tex/tab_editing_comparison}
\subsection{Applications}
\label{application}
Our Qffusion can deliver three applications: portrait video editing, whole-body driving~\cite{animate}, and jump cut smooth~\cite{jumpcut}.

\noindent\textbf{Portrait video editing.}
\lmm{We compare Qffusion with the current state-of-the-art video editing methods Codef~\cite{codef}, Rerender-A-Video~\cite{yang2023rerender}, TokenFlow~\cite{tokenflow} and AnyV2V~\cite{anyv2v} in Fig.~\ref{fig:video_editing}}. The editing scenarios consist of modifying style and hair, and adding sunglasses. 
As a text-driven editing method, TokenFlow~\cite{tokenflow} and Rerender-A-Video~\cite{yang2023rerender} cannot deal with some fine-grained local editing, such as hair editing. 
Codef~\cite{codef} sometimes suffers from the inconsistent ID issue (left example). Note that Codef~\cite{codef} relies on generating a canonical image to record static content in a video, which, however, usually has artifacts. When we further edit it with the desired style or appearance, these artifacts will also be spread throughout the entire video. 
In contrast, our editing results are identity-consistent and abide by the conditional poses clearly.
Moreover, although AnyV2V~\cite{anyv2v} performs image-driven video editing, it faces the degradation of editing appearance. 
\mm{Besides, we use ControlNext-SVD~\cite{peng2024controlnext} for comparison, which integrates ControlNet~\cite{controlnet} into an I2V model SVD~\cite{svd} for video controllable video generation. Here, we use facial keypoints as driving conditions. However, it yields inferior performance to our Qffusion.} More importantly, none of the existing methods can generate arbitrary-long videos.

\input{table_tex/quantitative_supp}

\lmm{We also provide more examples of our Qffusion on portrait video editing in our Appendix.} 
To further evaluate the video editing ability of Qffusion, we provide a quantitative comparison in Tab.~\ref{tab:editing_compare}. Specifically, we report the average CLIP-Image similarity and Warp Error. Our method yields the best performance on CLIP score and Warp Error,  which demonstrates that Qffusion can achieve amazing reference alignment and motion consistency. \lmm{Note that we cannot calculate CLIP-Image for TokenFlow~\cite{tokenflow} and Rerender-A-Video~\cite{yang2023rerender}, since they are text-driven methods.}


\noindent\textbf{Whole-body driving.} In Fig.~\ref{fig:who_body}, we display our Qffusion can also perform whole-body animation, where UBCFasion~\cite{zablotskaia2019dwnet} dataset is applied.
For the conditions, we use DWPose~\cite{yang2023effective} to detect landmarks for the face and body.
Specifically, we provide the visual comparison with the current state-of-the-art methods: DreamPose~\cite{dreampose}, BDMM~\cite{bdmm}, and Animate Anyone~\cite{animate}.
We apply the re-produced version~\cite{animatemoore} for Animate Anyone since its official code and dataset are not publicly available.
These methods are either carefully designed for whole-body animation tasks (\ie, DreamPose, BDMM), or need an additional heavy appearance network to encode the appearance of input identities (\ie, Animate Anyone).
It is difficult for DreamPose to ensure clothing consistency. Besides, Animate Anyone struggles to guarantee facial fidelity.
Without additional modules, our method can obtain better results than these task-specific methods, demonstrating the generalizability of the proposed method. 

\input{fig_tex/jump_cut}

\input{fig_tex/user_study}
We also give the quantitative comparison in Tab.~\ref{tab:quantitative_supp}. It can be seen that our method can achieve better results than those task-specific state-of-the-art techniques. The corresponding video results are shown in our supplement.

\noindent\textbf{Jump Cut Smooth.} 
A jump cut brings an abrupt, sometimes unwanted change in the viewing experience. Our method can be used to smooth these jump cuts. Fig.~\ref{fig:jump_cut} presents an extra application that our Qffusion can deal with, i.e., jump cut smooth~\cite{jumpcut}.
The application is performed as follows. 1) We take the jump-cut start frame as $\mathbf{I}^s$, end frame as $\mathbf{I}^e$. 2) We extract these two frames' conditions $\mathbf{C}^s$ and $\mathbf{C}^e$. 3) Assuming there are $K$ frames to be generated, we interpolate intermediate conditions $\mathbf{C}^k_P = (k/K) \mathbf{C}^s_P + (1 - k/K)\mathbf{C}^e_P, k \in \{1, \dots, K\}$. Here $\mathbf{C}^*_P$ denotes the conditional 3D points and $\mathbf{C}^*$ is the visualization of $\mathbf{C}^*_P$. As seen in Fig.~\ref{fig:jump_cut}, our Qffusion can achieve seamless transitions between cuts, even in challenging cases where the talking head undergoes large-scale movement or rotation in the jump cut.

\input{fig_tex/limitation}

\subsection{User Study}
We also provide a user study to compare our method with recently proposed video editing methods TokenFlow~\cite{tokenflow}, Codef~\cite{codef}, and AnyV2V~\cite{anyv2v}.
Specifically, we pose questions Q1: General Editing (GE), Q2: Temporal Consistency (TC), and Q3: Video Quality (VQ) to 30 anonymous participants on a crowd-sourcing platform, for randomly selected 12 video editing samples. We report the select ratio of four video editing methods in Fig.~\ref{fig:user_study}. Our Qffusion earns the highest user preference in all three aspects.

\section{Limitations}
While our method provides promising results for various applications, there still exist some limitations. For instance, as seen in Fig.~\ref{fig:limitation}, Qffusion sometimes faces unsatisfying cross-ID portrait video animation results. The reason here is that when the driving landmarks come from a different person, the shape information cannot be well-aligned.

%% file: table_tex/quantitative.tex
\begin{table}[t]
\setlength\tabcolsep{0.6pt}
    \centering
    \small
    \begin{tabular}{l|c|c|c|c}
    \toprule 
    
      Method   & PSNR$\uparrow$ & SSIM$\uparrow$ & Warp Error$\downarrow$  & LPIPS$\downarrow$ \\
    \hline
     (a) {GAN} &17.64  &{0.548} &\textbf{0.654}  &0.350  \\
     (b) ControlNet &\underline{18.29} &0.544 &2.124 &{0.322} \\
     (c) ControlNet+AnimateDiff
    &10.69 &0.212 &20.28 &{0.636} \\
(d) \mm{Animate Anyone*} &\mm{17.87} &\mm{\underline{0.565}}  &\mm{4.802}  &\mm{\underline{0.224}}\\
(e) \lmm{MagicAnimate} & \lmm{13.68}  &\lmm{0.321} & \lmm{5.077} & \lmm{0.489} \\
\hline
 \rowcolor{mygray}
     (f) Ours &\textbf{25.22} &\textbf{0.834} &\underline{0.665 } &\textbf{0.154} \\
     \bottomrule
    \end{tabular}   
    \caption{\mm{Quantitative comparisons with state-of-the-art animation methods, where our method yields the best performance on PSNR, SSIM, LPIPS, and competitive Warp Error.}}
  \label{tab:quantitative}
\end{table}

%% file: table_tex/ablation.tex
\begin{table}[t]
\begin{minipage}{0.45\textwidth}
\centering
\includegraphics[width=\linewidth]{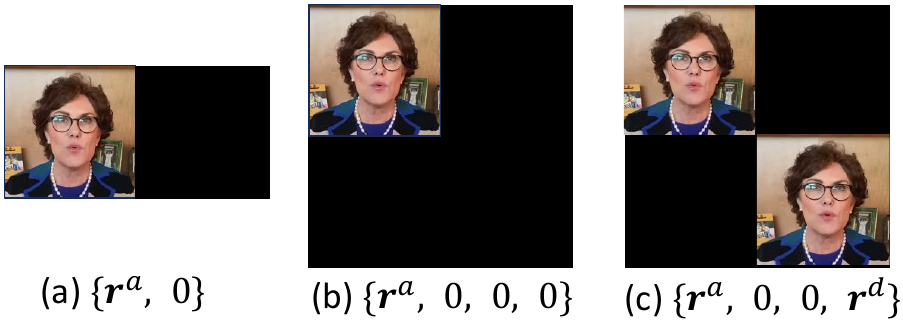}
\end{minipage}
\begin{minipage}{0.45\textwidth}
    \centering
  \renewcommand{\arraystretch}{1.4}
  \setlength\tabcolsep{1pt}
    \begin{tabular}{l|c|c|c|c}
    \toprule 
      Setting   & PSNR$\uparrow$ & SSIM$\uparrow$ & Warp Error$\downarrow$  & LPIPS$\downarrow$ \\
    \hline
     (a) $\{\mathbf{r^a}, \mathbf{0}\}$  &19.78  &0.648  &1.985  &0.272  \\
     (b) $\{\mathbf{r^a}, \mathbf{0},\mathbf{0},\mathbf{0}\}$ &23.87  &0.803  &1.760  &0.165\\
      \rowcolor{mygray}
     (c) Our $\{\mathbf{r^a}, \mathbf{0},\mathbf{0},\mathbf{r^d}\}$  &\textbf{25.22} &\textbf{0.834} &\textbf{0.665} &\textbf{0.154} \\
     \bottomrule
    \end{tabular}    
\end{minipage}
    \caption{Quantitative ablation studies on different settings of input arrangement.}
    \label{tab:ablation}
\end{table}

%% file: fig_tex/reconstruction.tex
\begin{figure*}[t]
    \centering
\includegraphics[width=1.0\linewidth]{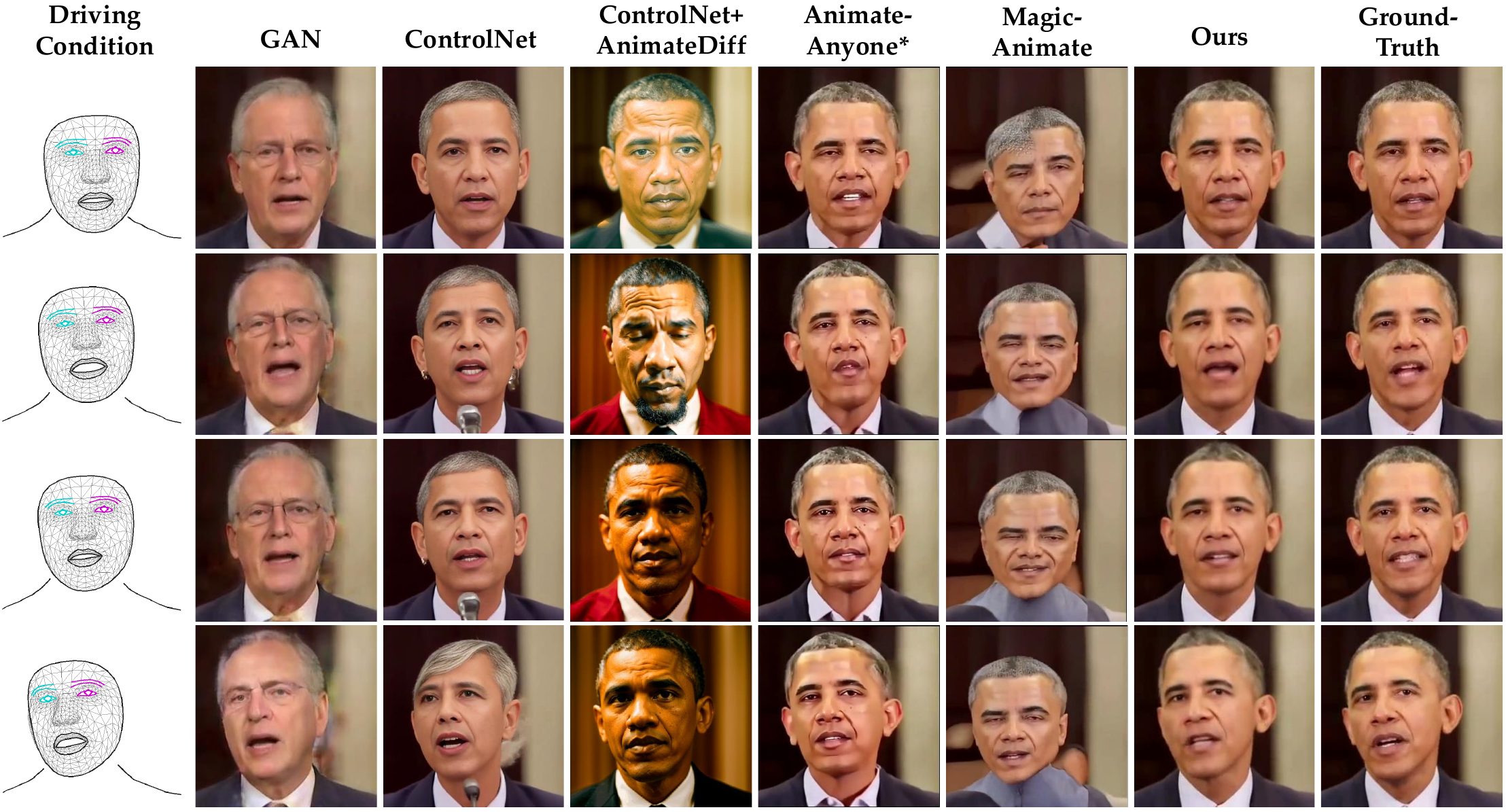}
    \caption{\mm{Animation comparisons with other methods.}}
    \label{fig:qualitative}
\end{figure*}

%% file: fig_tex/ablation.tex
\begin{figure}[t]
  \centering
  \includegraphics[width=1.0\linewidth]{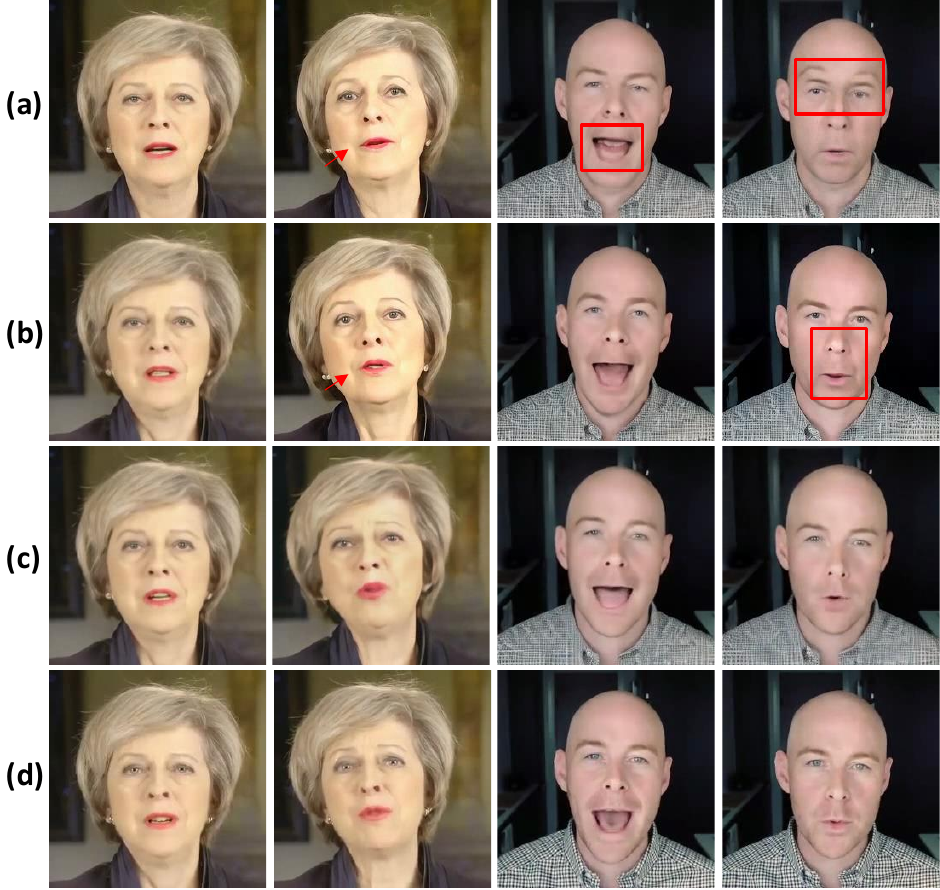} 
       \caption{
      Qualitative ablation studies of input arrangement. (a) two images  $\{\mathbf{r}^a, \mathbf{0}\}$; (b) a four-grid arrangement with one reference image  $\{\mathbf{r}^a, \mathbf{0},\mathbf{0},\mathbf{0}\}$ ; (c) our quadrant-grid design $\{\mathbf{r}^a, \mathbf{0},\mathbf{0},\mathbf{r}^d\}$; (d) ground-truth, respectively. All results are generated using the same training and inference strategies.
       }  
    \label{fig:ablation1}
\end{figure}


%% file: fig_tex/inference_supp.tex
\begin{figure}[t]
    \centering
    \includegraphics[width=1.0\linewidth]{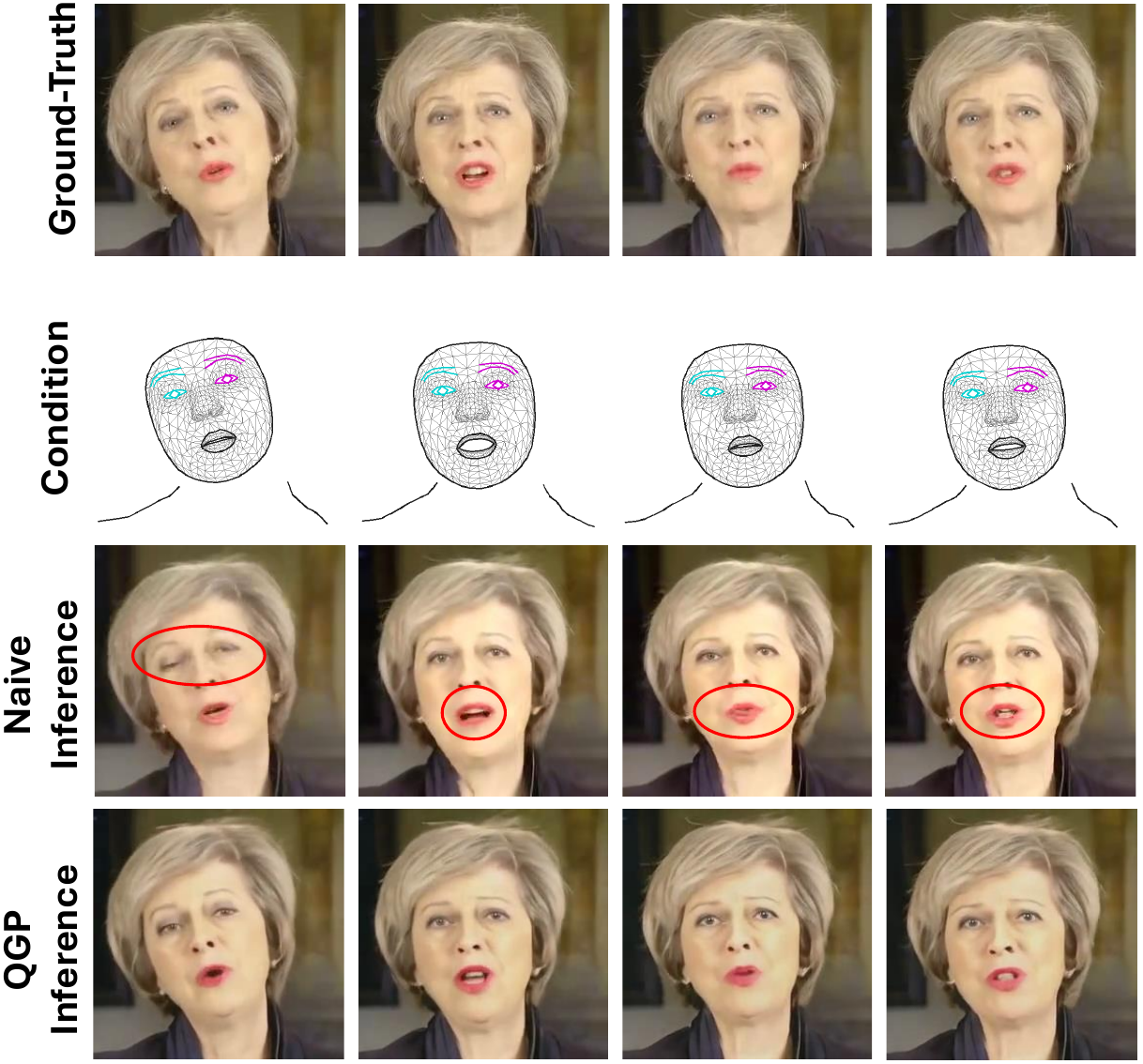}
    \caption{Qualitative ablation of our \texttt{QGP} inference and the naive inference, where \texttt{QGP} results are more motion-aligned and appearance-consistent.}  
    \label{fig:inference_supp}

\end{figure}

%% file: table_tex/tab_inference.tex
\begin{table}[t]
    \centering
    \small
    \begin{tabular}{l|c|c|c|c}
    \toprule  
      Method   & PSNR$\uparrow$ & SSIM$\uparrow$ & Warp Error$\downarrow$  & LPIPS$\downarrow$ \\
    \hline
     Naive Inference   &25.01  &{0.827} &{2.488}  &0.160  \\
\hline
 \rowcolor{mygray}
   Our \texttt{QGP} &\textbf{25.22} &\textbf{0.834} &\textbf{0.665 } &\textbf{0.154} \\
     \bottomrule
    \end{tabular}    
    \caption{Quantitative ablation studies of naive inference strategy and the proposed recursive \texttt{QGP} inference strategy.}
    \label{tab:inference}
\end{table}

%% file: fig_tex/video_editing.tex
\begin{figure*}[t]
    \centering
\includegraphics[width=1\linewidth]{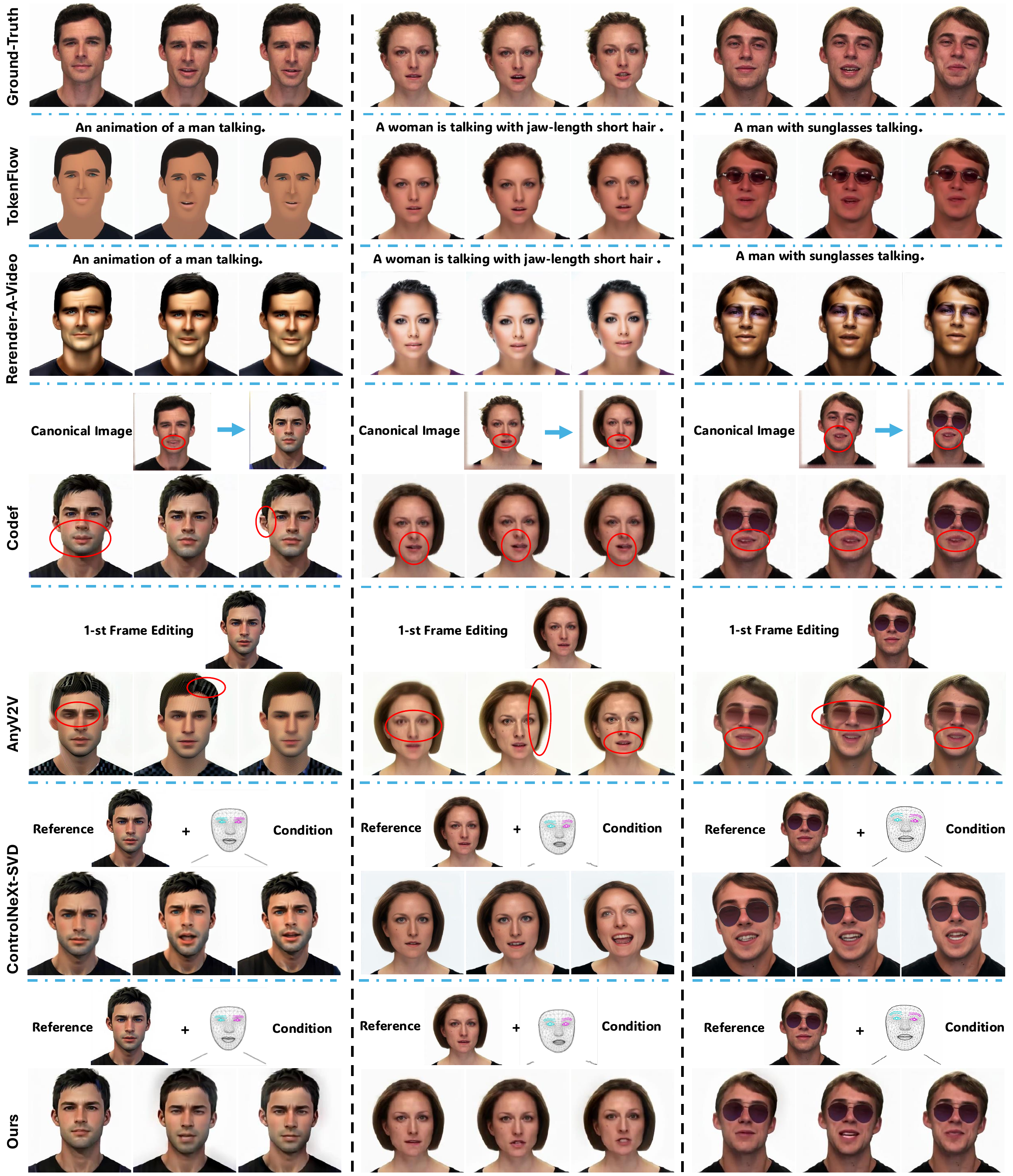}
    \caption{\lmm{We compare Qffusion with four recent video editing methods: TokenFlow~\cite{tokenflow}, Rerender-A-Video~\cite{yang2023rerender}, Codef~\cite{codef} and AnyV2V~\cite{anyv2v} on portrait video editing.} Besides, we provide animation results from ControlNext-SVD~\cite{peng2024controlnext}. Note that our method requires both modified start and end frames ($\mathbf{I}^s$ and $\mathbf{I}^e$) as editing signals, where $\mathbf{I}^e$ is omitted here for simplicity.}
    \label{fig:video_editing}
\end{figure*}



%% file: fig_tex/whole_body.tex
\begin{figure*}[t]
    \centering
\includegraphics[width=1\linewidth]{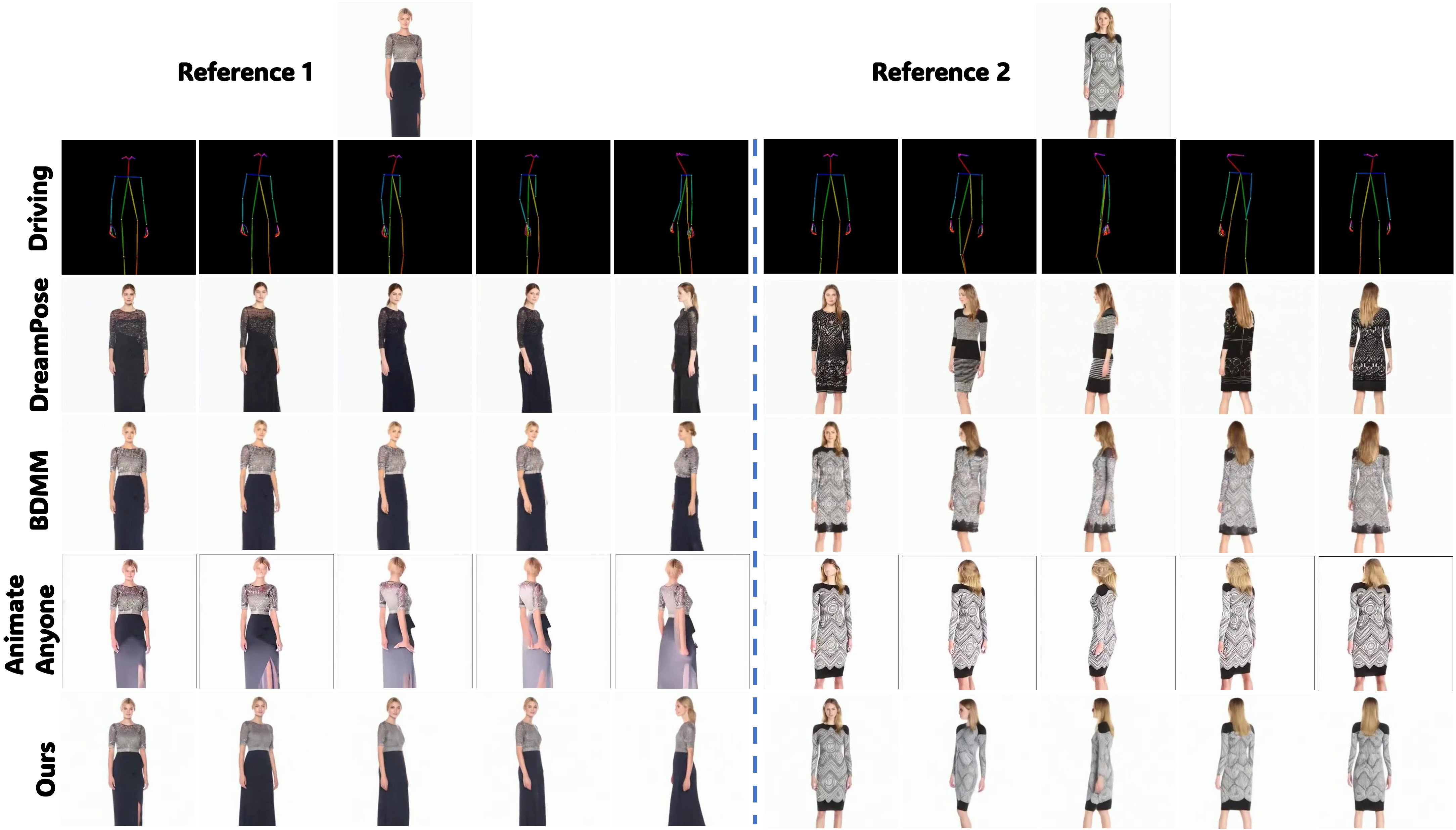}
    \caption{Whole-body driving results. We compare our Qffusion with  DreamPose~\cite{dreampose}, BDMM~\cite{bdmm}, Animate Anyone~\cite{animate}. We achieve better results than those task-specific methods, where the end reference is omitted for simplicity.
    }
     \vspace{-1mm}
    \label{fig:who_body}
\end{figure*}

%% file: table_tex/tab_editing_comparison.tex
\begin{table}[t]
\setlength\tabcolsep{7pt}
    \centering
    \small
    \begin{tabular}{l|c|c}
    \toprule  
      Method   & CLIP-Image $\uparrow$    & Warp Error$\downarrow$ \\
    \hline
   TokenFlow~\cite{tokenflow}   & - & 0.874 \\
\hline
\lmm{Rerender-A-Video~\cite{yang2023rerender}}  & - & \lmm{0.759} \\
\hline
   Codef~\cite{codef} &{0.938}  &0.301  \\
\hline
  \mm{ControlNext-SVD}~\cite{peng2024controlnext} &\mm{0.875} &\mm{3.606}  \\
 \hline
    AnyV2V~\cite{anyv2v} &0.924& 0.822 \\
 \hline
  \rowcolor{mygray}
   Ours & \textbf{0.959}  & \textbf{0.206} \\
     \bottomrule
    \end{tabular}    
    \caption{\mm{Quantitative comparison with other video editing techniques on CLIP-Image similarity and Warp Error. Besides, we provide the performance of ControlNext-SVD~\cite{peng2024controlnext}.
    }}
     \vspace{-1mm}
\label{tab:editing_compare}
\end{table}


%% file: table_tex/quantitative_supp.tex
\begin{table}[t]
\setlength\tabcolsep{3pt}
    \centering
    \small
    \begin{tabular}{l|c|c|c|c}
    \toprule 
    
      Method   & PSNR$\uparrow$ & SSIM$\uparrow$ & Warp Error$\downarrow$  & LPIPS$\downarrow$ \\
    \hline
     DreamPose~\cite{dreampose} & 15.21 &0.803 &4.688  &0.165  \\
     BDMM~\cite{bdmm} &22.26 &0.853 &1.516 &0.094 \\
 Animate Anyone~\cite{animate}
    &17.23 &0.762 &7.599 & 0.206\\
\hline
 \rowcolor{mygray}
Ours &\textbf{23.42} &\textbf{0.856} &\textbf{0.901} &\textbf{0.092} \\
     \bottomrule
    \end{tabular}   
    \vspace{-2mm}
    \caption{Quantitative comparisons with the current sate-of-the-art whole-body driving methods, where our method yields the best performance on PSNR, SSIM, LPIPS, and Warp Error.}
    \label{tab:quantitative_supp}
\end{table}

%% file: fig_tex/jump_cut.tex
\begin{figure*}[t]
    \centering
\includegraphics[width=0.96\linewidth]{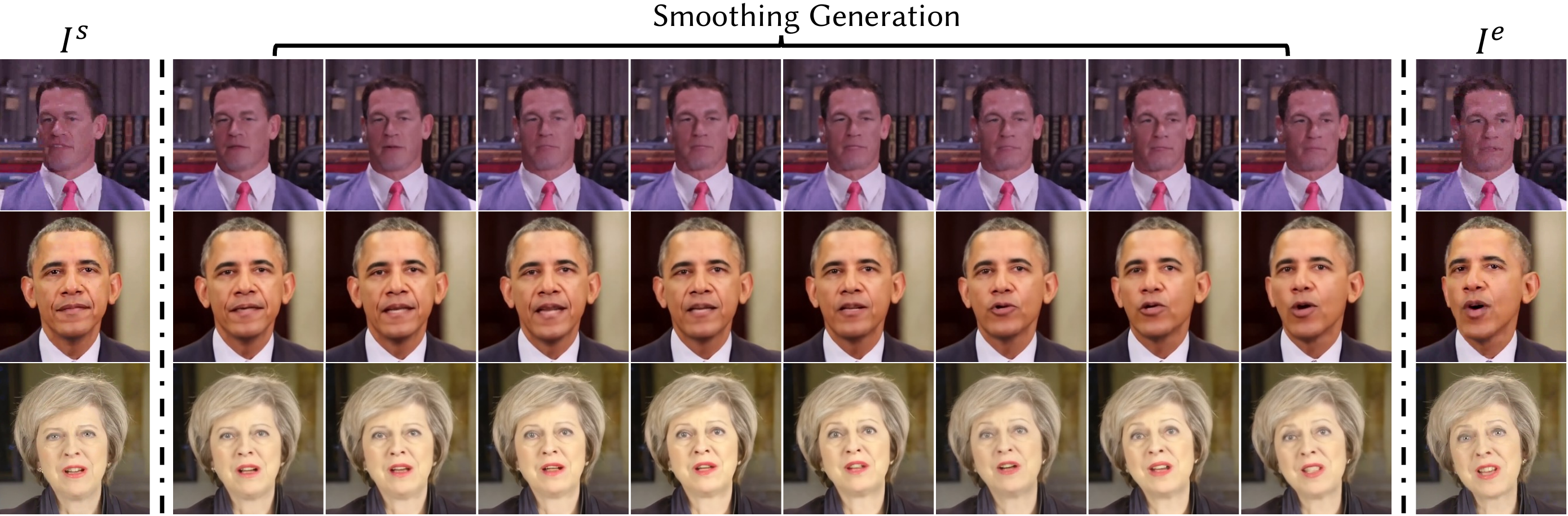}
    \caption{Jump cut smooth~\cite{jumpcut} results using our Qffusion. It has potential value for speech video editing and the film industry.}
    \label{fig:jump_cut}
\end{figure*}

%% file: fig_tex/user_study.tex
\begin{figure}[t]
  \centering
    \includegraphics[width=0.9\linewidth]{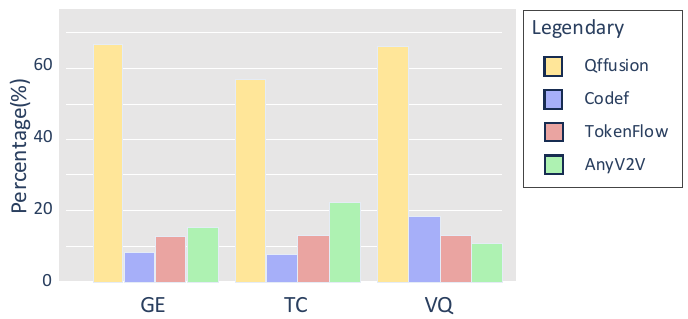}
   \caption{User study on the selected ratio. Our Qffusion outperforms the current state-of-the-art video editing methods in all three aspects.}
   \label{fig:user_study}
\end{figure}


%% file: fig_tex/limitation.tex
\begin{figure}[t]
    \centering
\includegraphics[width=1.0\linewidth]{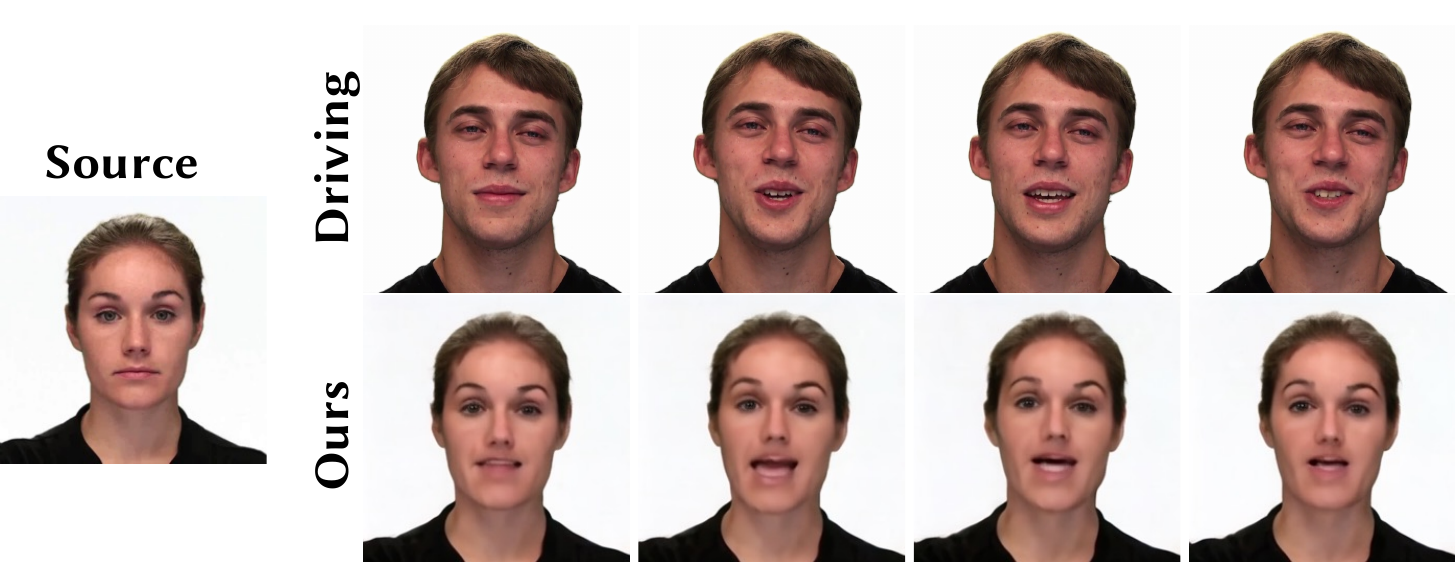}
    \caption{Our limitation. Our Qffusion struggles with ID fidelity when using different identities as reference images.}
    \label{fig:limitation}
\end{figure}

%% file: tex/conclusion.tex
\section{Conclusion}
This paper presented a \emph{dual-frame-guided} portrait video editing framework dubbed Qffusion, where we first modify the start and end frames with professional software (e.g., Photoshop or Meitu) and then propagate these modifications. Specifically, obeying an “animation for editing” principle, our Qffusion is trained as a general video animation model, which can be used for portrait video editing by treating the edited start and end frames as references. Specifically, we design a Quadrant-grid Arrangement (\texttt{QGA}) scheme for latent re-arrangement in SD, which captures spatial correspondence and temporal clues in a quadrant-grid design. Besides, stable arbitrary-length videos can be generated stably via our proposed recursive Quadrant-grid Propagation (\texttt{QGP}) inference. Our Qffusion serves as a foundational method, with the potential for future extension into various applications.

%% file: sec/X_suppl.tex
\clearpage
\setcounter{page}{1}
\setcounter{figure}{0}
\setcounter{table}{0}
\renewcommand{\figurename}{\textbf{Ap-Fig.}}
\renewcommand{\tablename}{\textbf{Ap-Tab.}}
\maketitlesupplementary

\appendix 
\lmm{To ensure the reproducibility and completeness of this paper, we make this Appendix with 4 sections. Appendix~\ref{a} provides more details for our \texttt{QGP} inference pipeline. Appendix~\ref{b} presents more of our portrait video animation capacity. Appendix~\ref{c} shows that we can perform arbitrary-long video animation and editing. Appendix~\ref{d} shows more results on portrait video editing.}

\section{More Details for QGP Inference}
\label{a}
To seek an even motion modeling, we propose a recursive influence strategy Quadrant-grid Propagation (\texttt{QGP}), in which we use generated frames as reference frames for next inference iteration.
Specifically, given a start frame $\mathbf{I}^s$, an end frame $\mathbf{I}^e$, and a sequence of conditions, this method takes the conditions as driving signals to recursively generate intermediate frames. The pseudo-code of the proposed inference pipeline is presented at Algorithm~\ref{alg:infer}, which mainly involves 4 steps:
\begin{itemize}
\item Given the two reference images $\mathbf{I}^i$ and $\mathbf{I}^{i+k}$ ($i$, $k$ indicate the start and the interval between start and ending frame, \ie, for the first step, $i=0,k=K$), we first generate the two intermediate images $\tilde{\mathbf{I}}^{\frac{k}{3}}, \tilde{\mathbf{I}}^{\frac{2k}{3}}$, where the index in Quadrant-grid Arrangement is 
$\left[0, \left \lfloor \frac{K}{3}\right \rfloor, \left \lfloor \frac{2K}{3}\right \rfloor, K \right]$. 
This makes the intervals between each sub-image in the quadrant the same, leading to a consistent temporal sampling.

\item Then we repeat step (i) to generate frames between 
$\left[0, \left \lfloor \frac{K}{3}\right \rfloor \right]$,
$\left[\left \lfloor \frac{K}{3}\right \rfloor, \left \lfloor \frac{2K}{3}\right \rfloor \right]$, and 
$\left[\left \lfloor \frac{2K}{3}\right \rfloor, K \right]$
following a consistent interval $\left \lfloor \frac{k}{9}\right \rfloor$.

\item We repeat step (ii) for the overall sequence. This process effectively establishes the relationship between the generated video frames and their preceding and succeeding frames.

\item We organize the intermediate video frames to form a newly generated video.
\end{itemize}

\begin{center}
\begin{algorithm}[h]
\caption{Quadrant-grid Propagation of Qffusion} \label{alg:infer}
\leftline{\textbf{Input}:} 
\leftline{\quad\quad Reference images $\mathbf{I}^0, \mathbf{I}^K$, Conditions $\{\mathbf{C}^0, \dots, \mathbf{C}^K\}$}
\leftline{\textbf{Output}:}
\leftline{\quad\quad Generated new video $\{\tilde{\mathbf{I}}^0, \dots, \tilde{\mathbf{I}}^K\}$}
\begin{algorithmic}[1]
    \State Queue = [[0, K]] \ \ \#[start index, interval between start and end]
    \While{Queue is not empty}
        \State $i, k$ = Queue[0]
        \State Queue.popleft()
        \State $\mathcal{Q}_0$ = \texttt{QGA}
         ($\mathbf{I}^i,\mathbf{C}^i, \mathbf{C}^{i+\frac{k}{3}}, \mathbf{C}^{i+\frac{2k}{3}}, \mathbf{C}^{i+k}, \mathbf{I}^{i+k}$) 
        \State $\tilde{\mathcal{Q}}_0$ = \texttt{Diffusion\&Denoising}($\mathcal{Q}_0$)
        \State $\tilde{\mathbf{z}}^{i+\frac{k}{3}}, \tilde{\mathbf{z}}^{i+\frac{2k}{3}} = \texttt{Split\&Unstack} (\tilde{\mathcal{Q}}_0)$ 
        \State $\tilde{\mathbf{I}}^{i+\frac{k}{3}}, \tilde{\mathbf{I}}^{i+\frac{2k}{3}} = \texttt{Split}(\mathcal{D}([\tilde{\mathbf{z}}^{i+\frac{k}{3}}; \tilde{\mathbf{z}}^{i+\frac{2k}{3}}]))$

        \For {j in [0, 1, 2]}
            \If{$i + k > K$}
                \State break
            \EndIf
            \State Queue.append($[i + \frac{k}{3} * j, \frac{k}{3}]$) \ \ \#append processed index
        \EndFor
    \EndWhile
    \State \textbf{return} $\{\tilde{\mathbf{I}}^0, \dots, \tilde{\mathbf{I}}^{K}\}$
\end{algorithmic}
\end{algorithm}
\end{center}

\input{fig_tex/supp_reconstruction}
\input{fig_tex/live_comp}

\input{fig_tex/long_video}
\input{fig_tex/long_video2}

We give an example of our \texttt{QGP} inference strategy. Assuming there is an 82-frame video indexed as $\{0-th, 1-st, ..., 80-th, 81-st\}$ to be generated. Then, in the first iteration, we apply the $0-th$ and $81-st$ frames as references to generate two intermediate frames: $27-th$ and $54-th$. Next, in the second iteration, we use the newly generated $27-th$ and $54-th$ as references. Concretely, we make $\{0-th, 27-th\}$, $\{27-th, 54-th\}$ and $\{54-th, 81-st\}$ as references separately to synthesize new intermediate frames, forming $\{0-th, 9-th, 18-th, 27-th\}$, $\{27-th, 36-th, 45-th, 54-th\}$, and $\{54-th, 63-th, 72-nd, 81-st\}$. The process would be stopped until all frames are produced. Note that our \texttt{QGP} inference gradually uses generated frames at the current inference iteration (e.g., $27-th$ and $54-th$) as reference frames for the next iteration, making the intervals between each sub-image in the four-grid representation the same in each iteration.

In contrast, the naive inference method fixes the start and end frames ($0-th$ and $81-st$) as references to gradually generate all intermediate frames $\{\{1-st, 80-th\}, \{2-nd, 79-th\}, ...\}$.

\input{fig_tex/old_video_editing}
\input{fig_tex/diff_ref}
\input{fig_tex/metric_comp}

\section{More Results on Portrait Video Animation}
\label{b}

Recall that in Fig. 4 of our main paper, we provide a qualitative comparison with different video animation methods, including: 
(i) {\textbf{GAN}}~\cite{lu2021live}, (ii) \textbf{ControlNet}~\cite{controlnet}, (iii) {\textbf{ControlNet}} $+$ {\textbf{Animatediff}}~\cite{animatediff}, \mm{(iv) \textbf{Animate Anyone*}}~\cite{animatemoore}, and  (v) \textbf{MagicAnimate}~\cite{xu2024magicanimate}. Here, we provide more visual comparison in Ap-Fig. \ref{fig:supp_qualitative}. The results demonstrate that compared with these animation methods, our Qffusion performs the best portrait-video animation on ID consistency and condition alignment.

\lmm{In addition to the above explicit-landmark-based animation methods, recent LivePortrait~\cite{guo2024liveportrait} explores
\textbf{implicit-keypoint-based} non-diffusion framework. Besides, LivePortrait and our Qffusion have the following differences. (i) The former uses a two-stage and mixed image-video training strategy, while our Qffusion can be trained in an end-to-end manner. (ii) Our Qffusion can easily be extended to whole-body video, while LivePortrait cannot. As seen in Ap-Fig.~\ref{fig:live_comp}, we provide a qualitative and quantitative comparison with LivePortrait. Our Qffusion can achieve competitive or better results with LivePortrait.}


\section{Long Video Animation and Editing}
\label{c}
\lmm{As a recursive inference method, our Quadrant-grid Propagation (QGP) also accumulates errors over time. However, it is worth noting that compared with the naive inference method, our QGP alleviates the error-accumulation issue by making a more even temporal sampling.}

\lmm{For qualitative illustration, we provide a 502-frame animation and editing results using the naive and QGP inference in Ap-Fig.~\ref{fig:long_video}, which shows our capability of long-term stability. Besides, in Ap-Fig.~\ref{fig:long_video_metric}, we provide a quantitative comparison between the naive and QGP inferences using PSNR and SSIM. It can be seen that the naive inference method accumulates exaggerated errors for long-term videos. It first uses $0-th$ and $501-th$ frames as reference frames to generate two frames $\{1-st, 500-th\}$. Then it uses these two generated frames $\{1-st, 500-th\}$ as references to generate the $\{2-nd, 499-th\}$ frames. That is, the process is conducted iteratively, where the newly generated frame pair $\{l-th, (501 - l)-th\}$ is used as reference to synthesize the next frame pair $\{(l + 1)-th, (500 - l - 1)-th\}$, until all intermediate frames are synthesized. Due to error accumulation, the later the generated frame, the worse the quality. 
}

\lmm{In contrast, our QGP inference has the same interval between sub-images of each four-grid representation. Specifically, in the first iteration, we apply the $0-th$ and $501-st$ frames as references to generate two intermediate frames: $167-th$ and $334-th$. Next, in the second iteration, we make $\{0-th, 167-th\}$, $\{167-th, 334-th\}$ and $\{334-th, 501-st\}$ as references separately to synthesize new intermediate frames. The process would be stopped until all frames are produced. From Ap-Fig.~\ref{fig:long_video_metric}, we find that our QGP inference can greatly alleviate error accumulation even in long-term video generation.}

\section{More Results on Portrait Video Editing}
\label{d}
To demonstrate that our Qffusion can generate videos aligning with the given conditions smoothly, we also provide more portrait video editing examples in Ap-Fig.~\ref{fig:old_video_editing}, where more driving conditions are presented.

\mm{Besides, one may wonder whether our method can deal with potential inconsistencies between the edited first and last frames. We give two sets of examples in Ap-Fig.~\ref{fig:diff_ref}, in which the edited start and end frames are slightly inconsistent and completely inconsistent, respectively. Our Qffusion can present smooth transition results. Besides, in Ap-Fig.~\ref{fig:metric_comp}, we quantitatively measure {CLIP-Image} similarity and Warp Error of editing video under three cases: "completely consistent", "slightly inconsistent" and "completely inconsistent". Here, {CLIP-Image} score is calculated by the edited start and end video frames, respectively. We found that when the edited first and last frames are inconsistent, the generated video tends to quickly align with the appearance of the end frame, which makes the {CLIP-Image} score of the edited end frame higher than that of the edited start frame. Besides, Warp Error gradually increases when the edited start and end frames are gradually inconsistent.}


%% file: fig_tex/supp_reconstruction.tex
\begin{figure*}[htbp]
    \centering
    \includegraphics[width=1.0\linewidth]{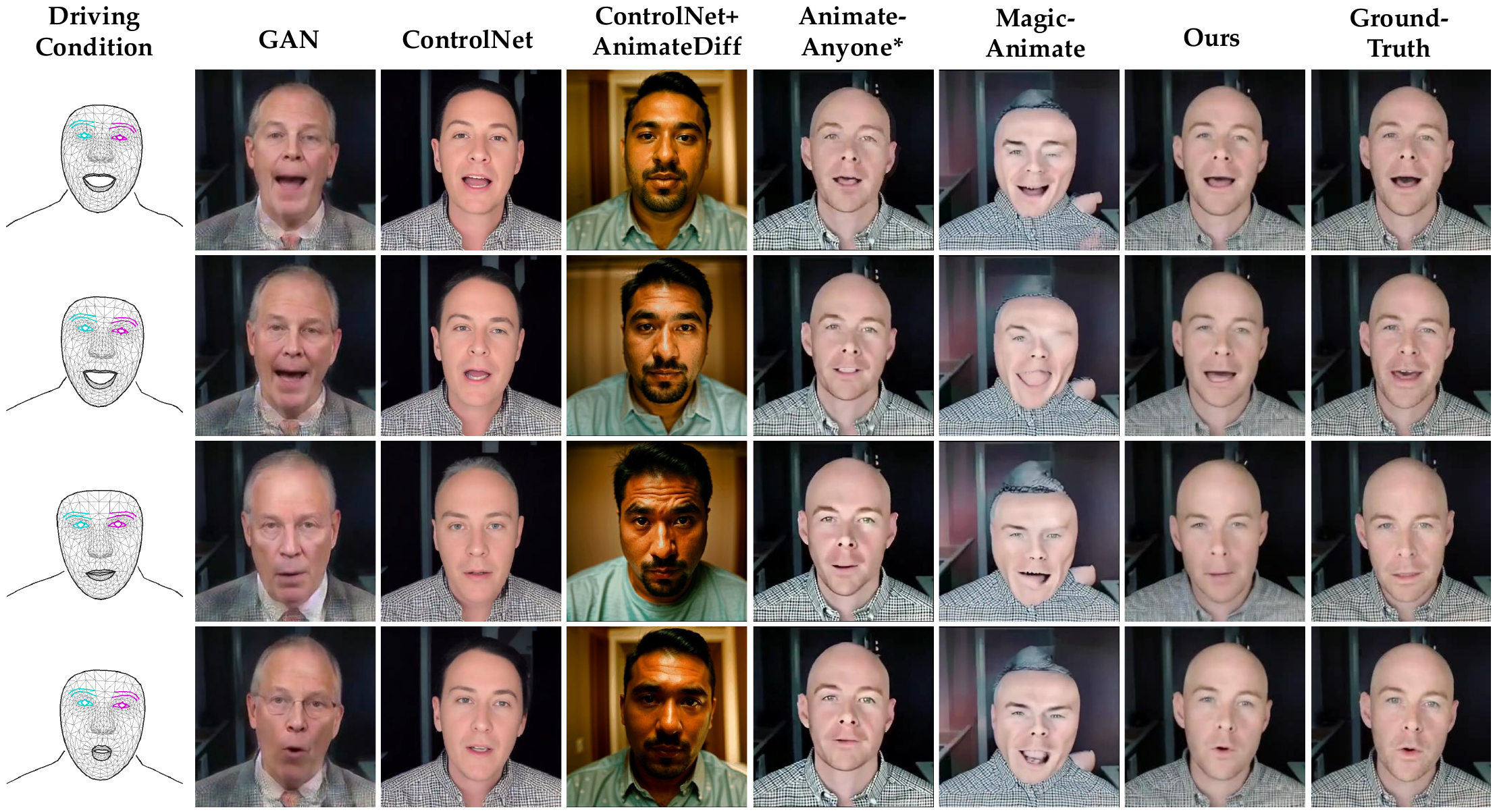}
    \caption{\lmm{More qualitative comparisons with other video animation methods using explicit landmarks.}}
    \label{fig:supp_qualitative}
\end{figure*}

%% file: fig_tex/live_comp.tex
\begin{figure*}[htbp]
  \centering
  \includegraphics[width=1.0\linewidth]{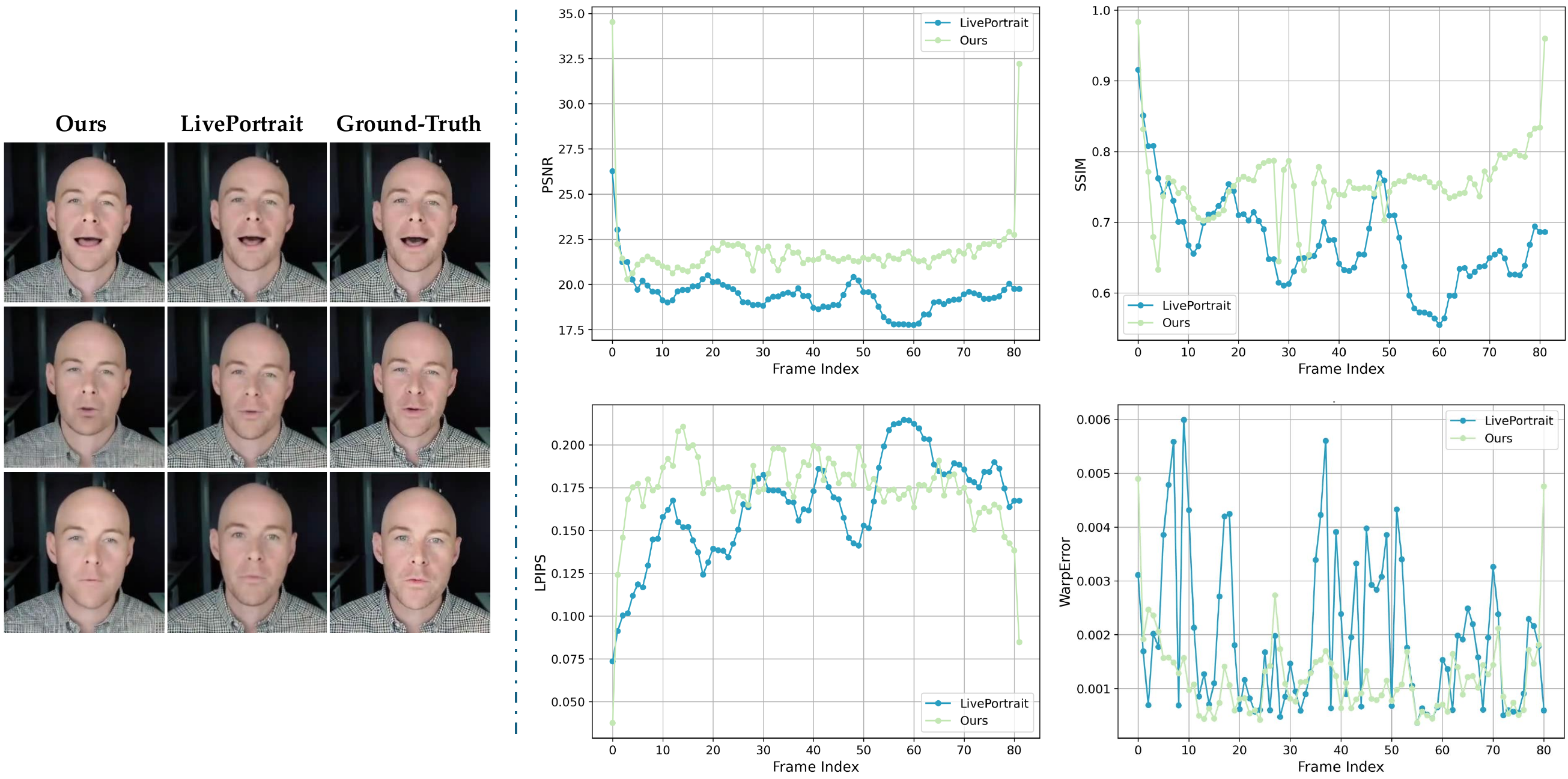} 
   \caption{\lmm{Qualitative and Quantitative comparisons with LivePortrait~\cite{guo2024liveportrait}, which is a non-diffusion animation method with \textbf{implicit} keypoints rather than explicit landmarks. Our Qffusion can achieve competitive or better results with LivePortrait.
       }}
    \label{fig:live_comp}
\end{figure*}

%% file: fig_tex/long_video.tex
\begin{figure*}[htbp]
    \centering
\includegraphics[width=1.0\linewidth]{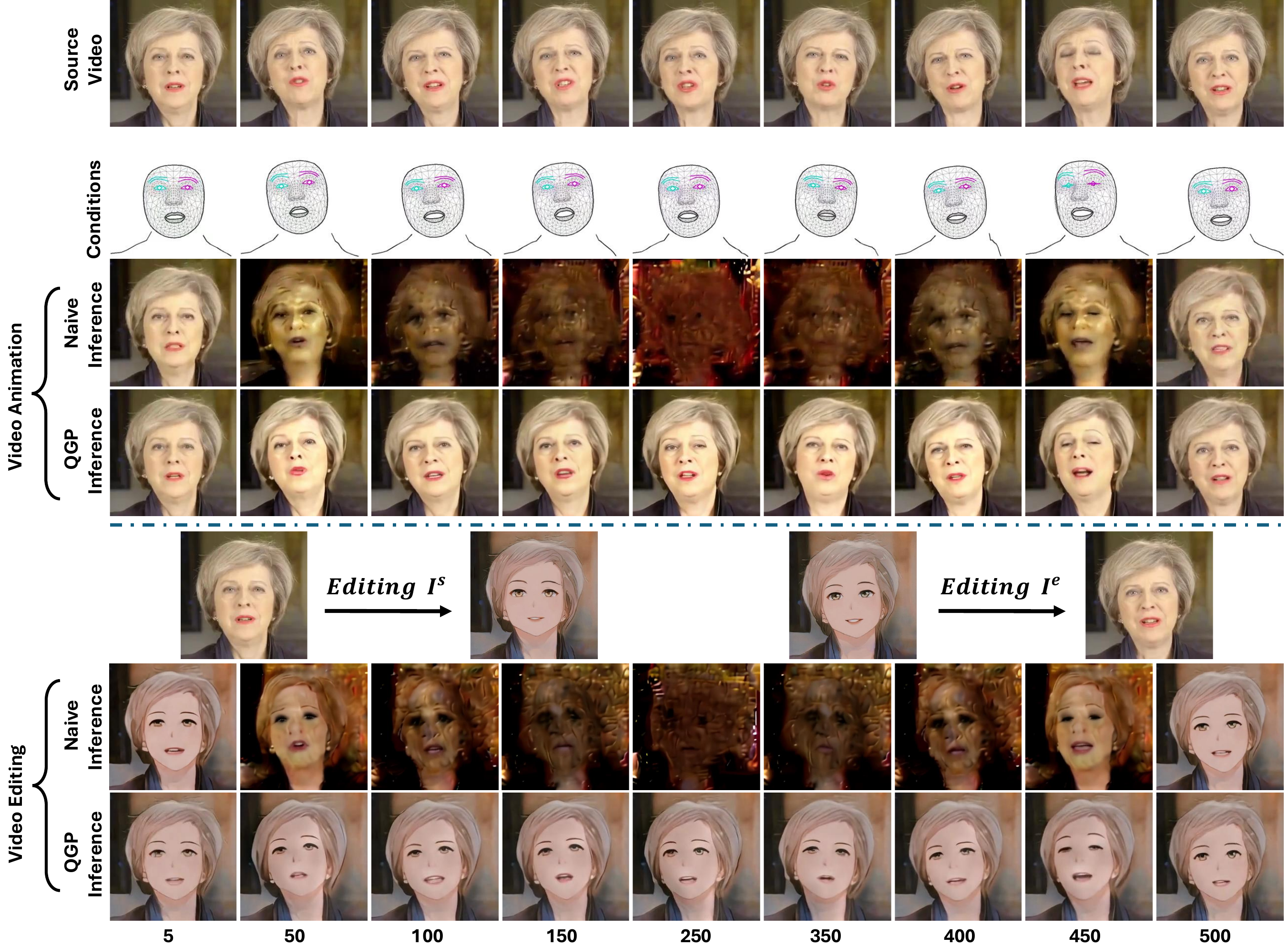}
\vspace{-2mm}
    \caption{\lmm{Qffusion can generate arbitrary-long videos while maintaining appearance and motion consistency, where we show 502-frame results. Top: using the naive and QGP inference to perform long-term video animation. Bottom: using the naive and QGP inference to perform long-term video editing.}}
    \label{fig:long_video}
\end{figure*}

%% file: fig_tex/long_video2.tex
\begin{figure*}[htbp]
    \centering
\includegraphics[width=1\linewidth]{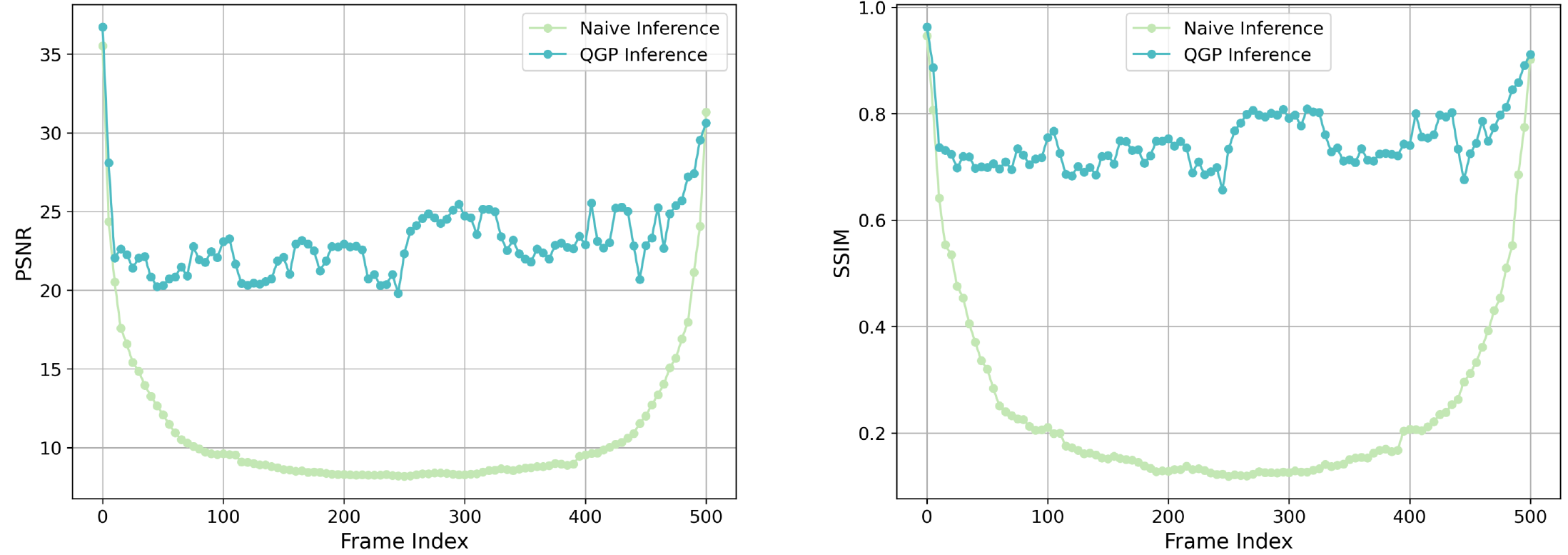}
\vspace{-2mm}
    \caption{\lmm{Quantitative comparison between the naive and QGP inference for long-term video generation. Here, we use PSNR and SSIM for illustration.}}
    \label{fig:long_video_metric}
\end{figure*}

%% file: fig_tex/old_video_editing.tex
\begin{figure*}[htbp]
    \centering
\includegraphics[width=1.0\linewidth]{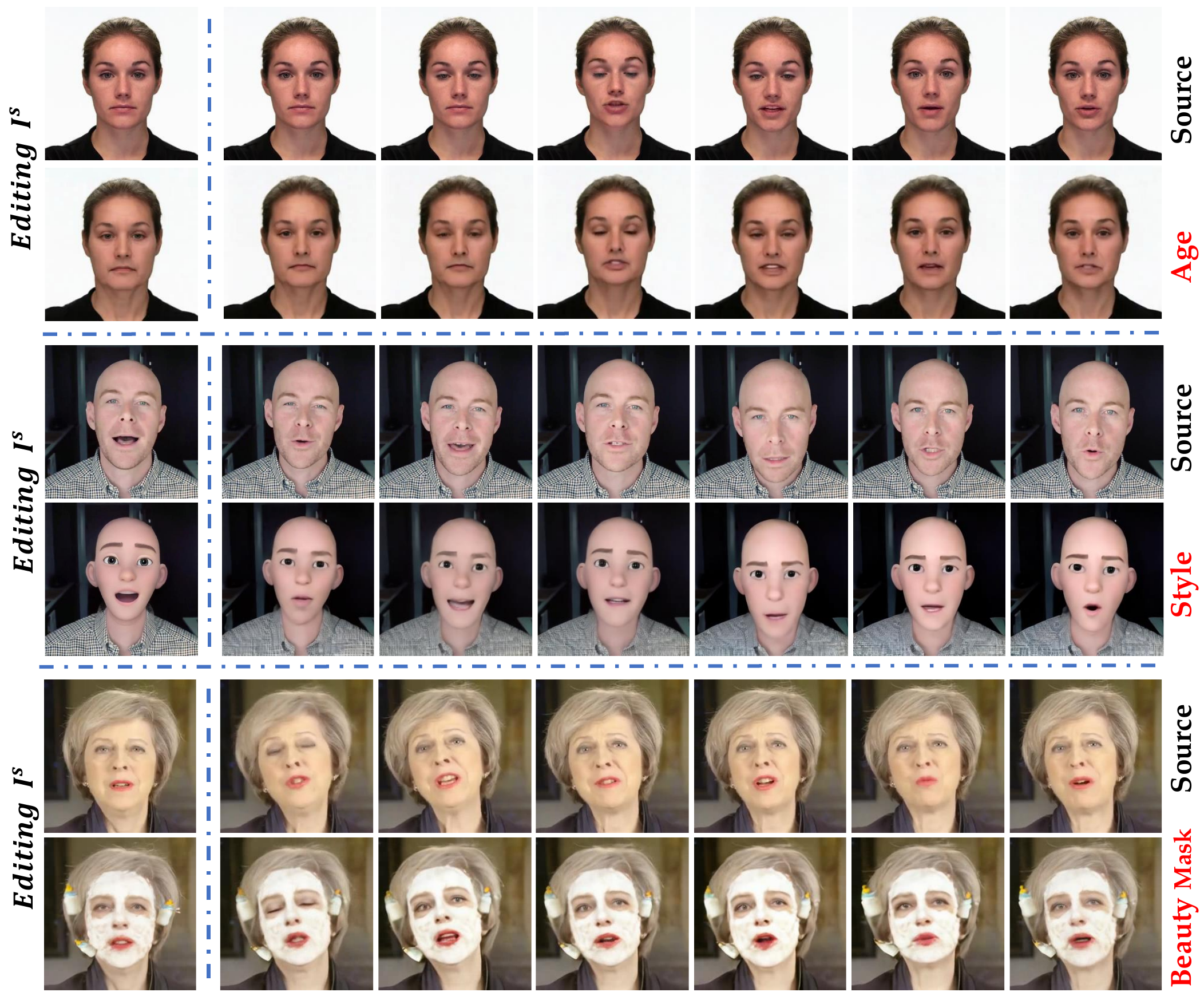}
    \caption{Our Qffusion can realize fine-grained local editing, including changing age and style, and adding  beauty masks. Note that Qffusion needs the modified start and end frame ($\mathbf{I}^s$ and $\mathbf{I}^e$) for editing propagation, where $\mathbf{I}^e$ is omitted for simplicity.
   } 
    \label{fig:old_video_editing}
\end{figure*}

%% file: fig_tex/diff_ref.tex
    \begin{figure*}[htbp]
      \centering
        \includegraphics[width=1.0\linewidth]{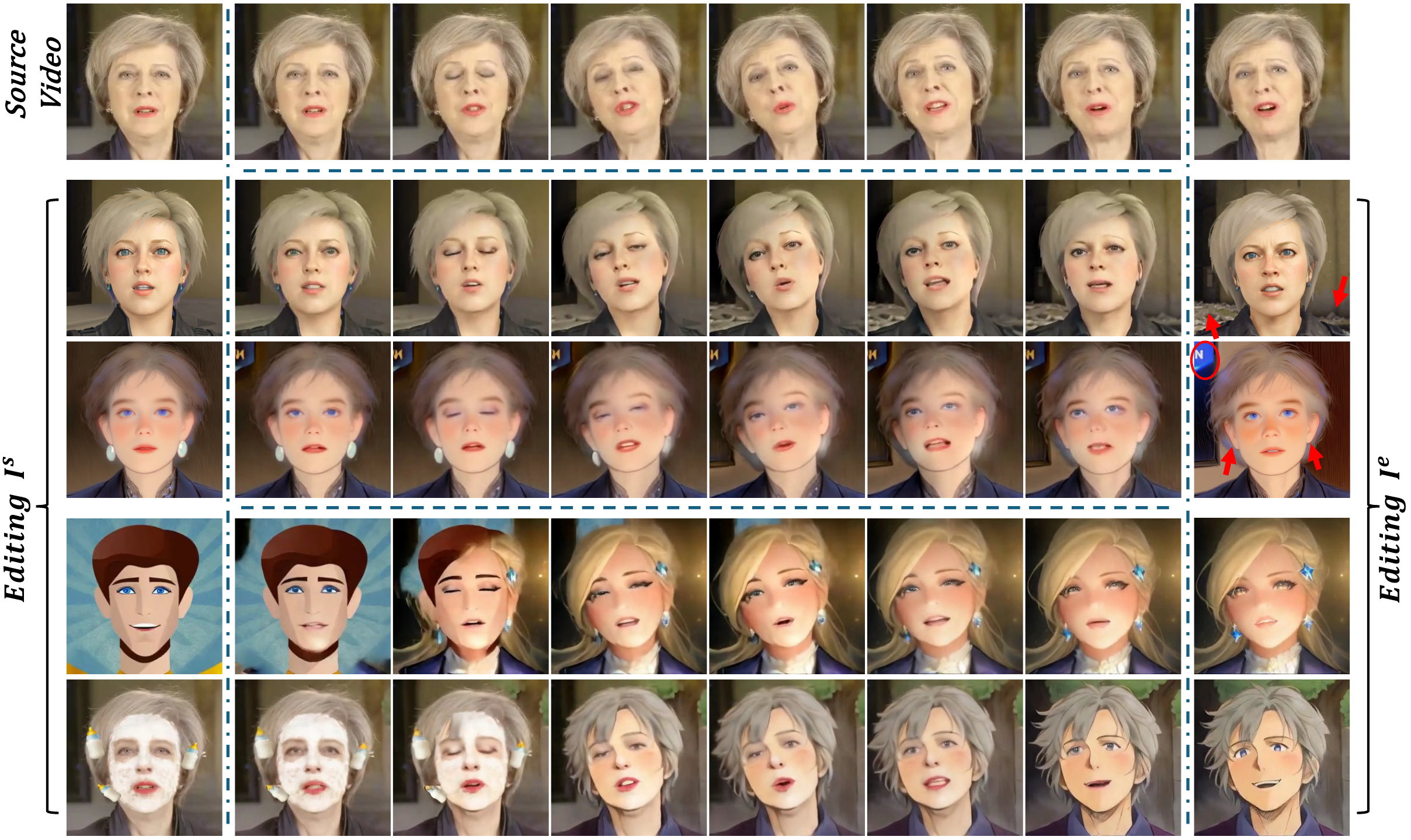}
       \caption{\mm{When the edited start and end frames are inconsistent, our method can present interesting transition results. We give two sets of examples, in which the start and end frames are slightly inconsistent and completely inconsistent, respectively.
}}
       \label{fig:diff_ref}
    \end{figure*}

%% file: fig_tex/metric_comp.tex
\begin{figure}[htbp]
    \centering
\includegraphics[width=1\linewidth]{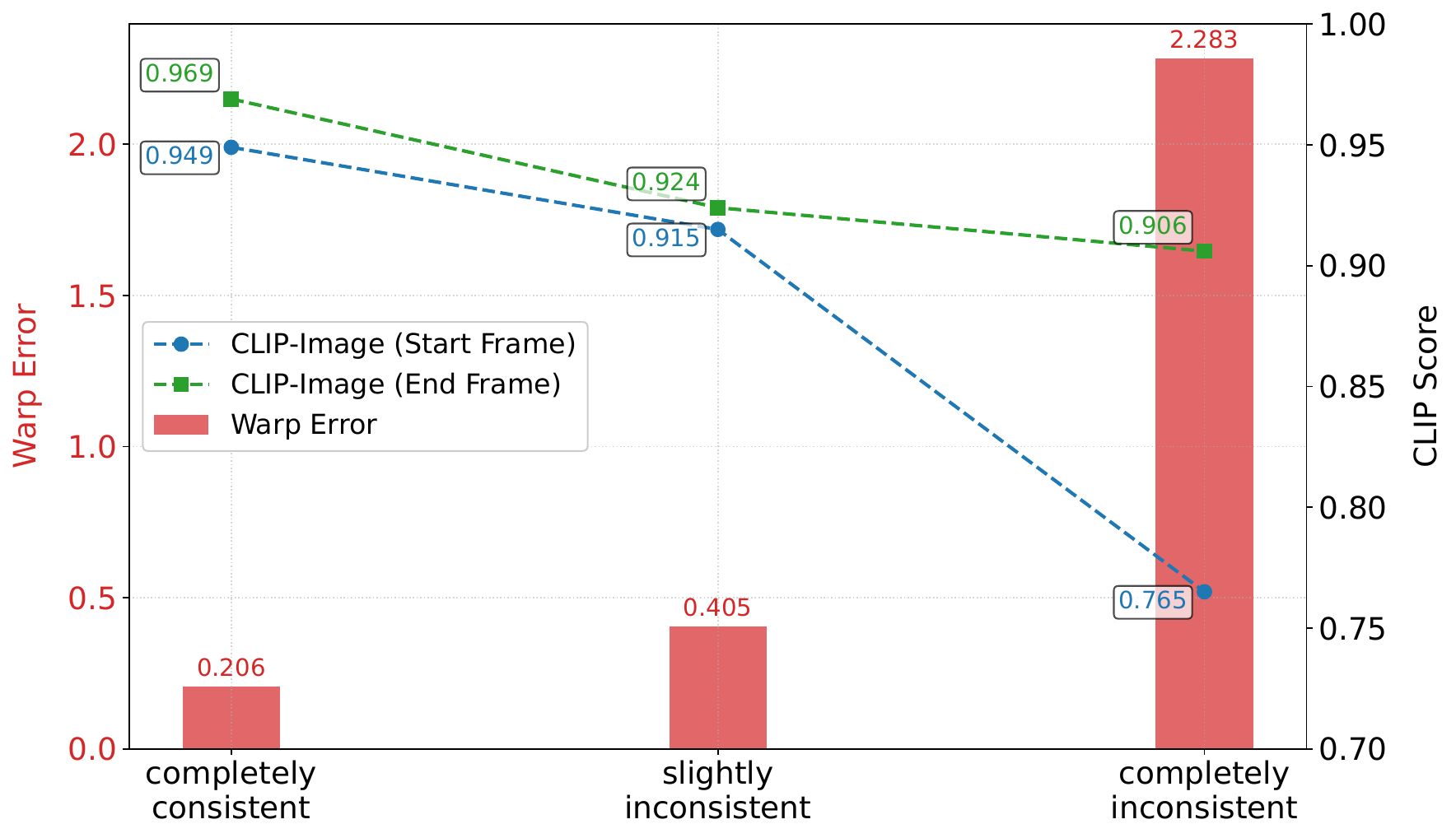}
\vspace{-4mm}
    \caption{\mm{Warp Error and CLIP-Image similarity across consistency conditions. Here, we calculate CLIP-Image similarity from the edited start frame and end frame, respectively.
}}
    \label{fig:metric_comp}
\end{figure}

%% file: main.bbl
\begin{thebibliography}{76}
\providecommand{\natexlab}[1]{#1}
\providecommand{\url}[1]{\texttt{#1}}
\expandafter\ifx\csname urlstyle\endcsname\relax
  \providecommand{\doi}[1]{doi: #1}\else
  \providecommand{\doi}{doi: \begingroup \urlstyle{rm}\Url}\fi

\bibitem[pik(2023)]{pika}
Pika labs.
\newblock \emph{URL https://pika.art/}, 2023.

\bibitem[gen(2024)]{gen-3}
Gen-3 alpha.
\newblock \emph{https://runwayml.com/research/introducing-gen-3-alpha}, 2024.

\bibitem[Avrahami et~al.(2023)Avrahami, Hayes, Gafni, Gupta, Taigman, Parikh,
  Lischinski, Fried, and Yin]{avrahami2023spatext}
Omri Avrahami, Thomas Hayes, Oran Gafni, Sonal Gupta, Yaniv Taigman, Devi
  Parikh, Dani Lischinski, Ohad Fried, and Xi Yin.
\newblock Spatext: Spatio-textual representation for controllable image
  generation.
\newblock In \emph{CVPR}, pages 18370--18380, 2023.

\bibitem[Bar-Tal et~al.(2022)Bar-Tal, Ofri-Amar, Fridman, Kasten, and
  Dekel]{2022text2live}
Omer Bar-Tal, Dolev Ofri-Amar, Rafail Fridman, Yoni Kasten, and Tali Dekel.
\newblock Text2live: Text-driven layered image and video editing.
\newblock In \emph{ECCV}, pages 707--723. Springer, 2022.

\bibitem[Blattmann et~al.(2023{\natexlab{a}})Blattmann, Dockhorn, Kulal,
  Mendelevitch, Kilian, Lorenz, Levi, English, Voleti, Letts, et~al.]{svd}
Andreas Blattmann, Tim Dockhorn, Sumith Kulal, Daniel Mendelevitch, Maciej
  Kilian, Dominik Lorenz, Yam Levi, Zion English, Vikram Voleti, Adam Letts,
  et~al.
\newblock Stable video diffusion: Scaling latent video diffusion models to
  large datasets.
\newblock \emph{arXiv preprint arXiv:2311.15127}, 2023{\natexlab{a}}.

\bibitem[Blattmann et~al.(2023{\natexlab{b}})Blattmann, Rombach, Ling,
  Dockhorn, Kim, Fidler, and Kreis]{blattmann2023align}
Andreas Blattmann, Robin Rombach, Huan Ling, Tim Dockhorn, Seung~Wook Kim,
  Sanja Fidler, and Karsten Kreis.
\newblock Align your latents: High-resolution video synthesis with latent
  diffusion models.
\newblock In \emph{CVPR}, pages 22563--22575, 2023{\natexlab{b}}.

\bibitem[Brooks et~al.(2023)Brooks, Holynski, and
  Efros]{brooks2023instructpix2pix}
Tim Brooks, Aleksander Holynski, and Alexei~A Efros.
\newblock Instructpix2pix: Learning to follow image editing instructions.
\newblock In \emph{CVPR}, pages 18392--18402, 2023.

\bibitem[Brooks et~al.(2024)Brooks, Peebles, Holmes, DePue, Guo, Jing, Schnurr,
  Taylor, Luhman, Luhman, et~al.]{sora}
Tim Brooks, Bill Peebles, Connor Holmes, Will DePue, Yufei Guo, Li Jing, David
  Schnurr, Joe Taylor, Troy Luhman, Eric Luhman, et~al.
\newblock Video generation models as world simulators. 2024.
\newblock \emph{URL https://openai.
  com/research/video-generation-models-as-world-simulators}, 3, 2024.

\bibitem[Burgert et~al.(2025)Burgert, Xu, Xian, Pilarski, Clausen, He, Ma,
  Deng, Li, Mousavi, et~al.]{burgert2025go}
Ryan Burgert, Yuancheng Xu, Wenqi Xian, Oliver Pilarski, Pascal Clausen,
  Mingming He, Li Ma, Yitong Deng, Lingxiao Li, Mohsen Mousavi, et~al.
\newblock Go-with-the-flow: Motion-controllable video diffusion models using
  real-time warped noise.
\newblock \emph{arXiv preprint arXiv:2501.08331}, 2025.

\bibitem[Cao et~al.(2023)Cao, Wang, Qi, Shan, Qie, and Zheng]{cao2023masactrl}
Mingdeng Cao, Xintao Wang, Zhongang Qi, Ying Shan, Xiaohu Qie, and Yinqiang
  Zheng.
\newblock Masactrl: Tuning-free mutual self-attention control for consistent
  image synthesis and editing.
\newblock In \emph{ICCV}, pages 22560--22570, 2023.

\bibitem[Ceylan et~al.(2023)Ceylan, Huang, and Mitra]{pix2video}
Duygu Ceylan, Chun-Hao~P Huang, and Niloy~J Mitra.
\newblock Pix2video: Video editing using image diffusion.
\newblock In \emph{ICCV}, pages 23206--23217, 2023.

\bibitem[Cong et~al.(2024)Cong, Xu, Simon, Chen, Ren, Xie, Perez-Rua,
  Rosenhahn, Xiang, and He]{cong2023flatten}
Yuren Cong, Mengmeng Xu, Christian Simon, Shoufa Chen, Jiawei Ren, Yanping Xie,
  Juan-Manuel Perez-Rua, Bodo Rosenhahn, Tao Xiang, and Sen He.
\newblock Flatten: Optical flow-guided attention for consistent text-to-video
  editing.
\newblock \emph{ICLR}, 2024.

\bibitem[Deng et~al.(2019)Deng, Guo, Xue, and Zafeiriou]{deng2019arcface}
Jiankang Deng, Jia Guo, Niannan Xue, and Stefanos Zafeiriou.
\newblock Arcface: Additive angular margin loss for deep face recognition.
\newblock In \emph{CVPR}, pages 4690--4699, 2019.

\bibitem[Dong et~al.(2023)Dong, Xue, Duan, and Han]{prompt}
Wenkai Dong, Song Xue, Xiaoyue Duan, and Shumin Han.
\newblock Prompt tuning inversion for text-driven image editing using diffusion
  models.
\newblock In \emph{ICCV}, pages 7430--7440, 2023.

\bibitem[Dong et~al.(2022)Dong, Wei, and Lin]{dreamartist}
Ziyi Dong, Pengxu Wei, and Liang Lin.
\newblock Dreamartist: Towards controllable one-shot text-to-image generation
  via contrastive prompt-tuning.
\newblock \emph{arXiv preprint arXiv:2211.11337}, 2022.

\bibitem[Esser et~al.(2023)Esser, Chiu, Atighehchian, Granskog, and
  Germanidis]{esser2023structure}
Patrick Esser, Johnathan Chiu, Parmida Atighehchian, Jonathan Granskog, and
  Anastasis Germanidis.
\newblock Structure and content-guided video synthesis with diffusion models.
\newblock In \emph{ICCV}, pages 7346--7356, 2023.

\bibitem[Gafni et~al.(2022)Gafni, Polyak, Ashual, Sheynin, Parikh, and
  Taigman]{gafni2022make}
Oran Gafni, Adam Polyak, Oron Ashual, Shelly Sheynin, Devi Parikh, and Yaniv
  Taigman.
\newblock Make-a-scene: Scene-based text-to-image generation with human priors.
\newblock In \emph{ECCV}, pages 89--106. Springer, 2022.

\bibitem[Gal et~al.(2022)Gal, Alaluf, Atzmon, Patashnik, Bermano, Chechik, and
  Cohen-Or]{textual-inversion}
Rinon Gal, Yuval Alaluf, Yuval Atzmon, Or Patashnik, Amit~H Bermano, Gal
  Chechik, and Daniel Cohen-Or.
\newblock An image is worth one word: Personalizing text-to-image generation
  using textual inversion.
\newblock \emph{arXiv preprint arXiv:2208.01618}, 2022.

\bibitem[Geyer et~al.(2024)Geyer, Bar-Tal, Bagon, and Dekel]{tokenflow}
Michal Geyer, Omer Bar-Tal, Shai Bagon, and Tali Dekel.
\newblock Tokenflow: Consistent diffusion features for consistent video
  editing.
\newblock \emph{ICLR}, 2024.

\bibitem[Goodfellow et~al.(2014)Goodfellow, Pouget-Abadie, Mirza, Xu,
  Warde-Farley, Ozair, Courville, and Bengio]{goodfellow2014generative}
Ian Goodfellow, Jean Pouget-Abadie, Mehdi Mirza, Bing Xu, David Warde-Farley,
  Sherjil Ozair, Aaron Courville, and Yoshua Bengio.
\newblock Generative adversarial nets.
\newblock \emph{NeurIPS}, 27, 2014.

\bibitem[Guo et~al.(2024{\natexlab{a}})Guo, Zhang, Liu, Zhong, Zhang, Wan, and
  Zhang]{guo2024liveportrait}
Jianzhu Guo, Dingyun Zhang, Xiaoqiang Liu, Zhizhou Zhong, Yuan Zhang, Pengfei
  Wan, and Di Zhang.
\newblock Liveportrait: Efficient portrait animation with stitching and
  retargeting control.
\newblock \emph{arXiv preprint arXiv:2407.03168}, 2024{\natexlab{a}}.

\bibitem[Guo et~al.(2024{\natexlab{b}})Guo, Yang, Rao, Wang, Qiao, Lin, and
  Dai]{animatediff}
Yuwei Guo, Ceyuan Yang, Anyi Rao, Yaohui Wang, Yu Qiao, Dahua Lin, and Bo Dai.
\newblock Animatediff: Animate your personalized text-to-image diffusion models
  without specific tuning.
\newblock \emph{ICLR}, 2024{\natexlab{b}}.

\bibitem[Hertz et~al.(2022)Hertz, Mokady, Tenenbaum, Aberman, Pritch, and
  Cohen-Or]{p2p}
Amir Hertz, Ron Mokady, Jay Tenenbaum, Kfir Aberman, Yael Pritch, and Daniel
  Cohen-Or.
\newblock Prompt-to-prompt image editing with cross attention control.
\newblock \emph{arXiv preprint arXiv:2208.01626}, 2022.

\bibitem[Ho et~al.(2020)Ho, Jain, and Abbeel]{ddpm}
Jonathan Ho, Ajay Jain, and Pieter Abbeel.
\newblock {Denoising diffusion probabilistic models}.
\newblock \emph{{NeurIPS}}, 33:\penalty0 6840--6851, 2020.

\bibitem[Ho et~al.(2022)Ho, Chan, Saharia, Whang, Gao, Gritsenko, Kingma,
  Poole, Norouzi, Fleet, et~al.]{imagen-video}
Jonathan Ho, William Chan, Chitwan Saharia, Jay Whang, Ruiqi Gao, Alexey
  Gritsenko, Diederik~P Kingma, Ben Poole, Mohammad Norouzi, David~J Fleet,
  et~al.
\newblock Imagen video: High definition video generation with diffusion models.
\newblock \emph{arXiv preprint arXiv:2210.02303}, 2022.

\bibitem[Hu(2024)]{animate}
Li Hu.
\newblock Animate anyone: Consistent and controllable image-to-video synthesis
  for character animation.
\newblock In \emph{CVPR}, pages 8153--8163, 2024.

\bibitem[Hu et~al.(2023)Hu, Gao, Zhang, Sun, Zhang, and Bo]{animatemoore}
Li Hu, Xin Gao, Peng Zhang, Ke Sun, Bang Zhang, and Liefeng Bo.
\newblock Animate anyone: Consistent and controllable image-to-video synthesis
  for character animation.
\newblock \url{https://github.com/MooreThreads/Moore-AnimateAnyone}, 2023.

\bibitem[Huynh-Thu and Ghanbari(2008)]{PSNR}
Quan Huynh-Thu and Mohammed Ghanbari.
\newblock Scope of validity of psnr in image/video quality assessment.
\newblock \emph{Electronics letters}, 44\penalty0 (13):\penalty0 800--801,
  2008.

\bibitem[Karras et~al.(2023)Karras, Holynski, Wang, and
  Kemelmacher-Shlizerman]{dreampose}
Johanna Karras, Aleksander Holynski, Ting-Chun Wang, and Ira
  Kemelmacher-Shlizerman.
\newblock Dreampose: Fashion image-to-video synthesis via stable diffusion.
\newblock \emph{arXiv preprint arXiv:2304.06025}, 2023.

\bibitem[Kingma and Welling(2013)]{vae}
Diederik~P Kingma and Max Welling.
\newblock Auto-encoding variational bayes.
\newblock \emph{arXiv preprint arXiv:1312.6114}, 2013.

\bibitem[Ku et~al.(2024)Ku, Wei, Ren, Yang, and Chen]{anyv2v}
Max Ku, Cong Wei, Weiming Ren, Huan Yang, and Wenhu Chen.
\newblock Anyv2v: A plug-and-play framework for any video-to-video editing
  tasks.
\newblock \emph{TMLR}, 2024.

\bibitem[Kumari et~al.(2023)Kumari, Zhang, Zhang, Shechtman, and
  Zhu]{custom-diffusion}
Nupur Kumari, Bingliang Zhang, Richard Zhang, Eli Shechtman, and Jun-Yan Zhu.
\newblock Multi-concept customization of text-to-image diffusion.
\newblock In \emph{ICCV}, pages 1931--1941, 2023.

\bibitem[Li et~al.(2024)Li, Li, Yang, Liu, Yue, Lin, and Xu]{li2023video}
Maomao Li, Yu Li, Tianyu Yang, Yunfei Liu, Dongxu Yue, Zhihui Lin, and Dong Xu.
\newblock A video is worth 256 bases: Spatial-temporal expectation-maximization
  inversion for zero-shot video editing.
\newblock In \emph{CVPR}, pages 7528--7537, 2024.

\bibitem[Liu et~al.(2023)Liu, Lin, Yu, Zhou, and Li]{liu2023moda}
Yunfei Liu, Lijian Lin, Fei Yu, Changyin Zhou, and Yu Li.
\newblock Moda: Mapping-once audio-driven portrait animation with dual
  attentions.
\newblock In \emph{ICCV}, pages 23020--23029, 2023.

\bibitem[Livingstone and Russo(2018)]{RAVDESS}
Steven~R Livingstone and Frank~A Russo.
\newblock The ryerson audio-visual database of emotional speech and song
  (ravdess): A dynamic, multimodal set of facial and vocal expressions in north
  american english.
\newblock \emph{PloS one}, 13\penalty0 (5):\penalty0 e0196391, 2018.

\bibitem[Loshchilov and Hutter(2017)]{Ilya2017AdamW}
Ilya Loshchilov and Frank Hutter.
\newblock Decoupled weight decay regularization.
\newblock In \emph{ICLR}, 2017.

\bibitem[Lu et~al.(2021)Lu, Chai, and Cao]{lu2021live}
Yuanxun Lu, Jinxiang Chai, and Xun Cao.
\newblock Live speech portraits: Real-time photorealistic talking-head
  animation.
\newblock \emph{ACM Transactions on Graphics (TOG)}, 40\penalty0 (6):\penalty0
  1--17, 2021.

\bibitem[Lugaresi et~al.(2019)Lugaresi, Tang, Nash, McClanahan, Uboweja, Hays,
  Zhang, Chang, Yong, Lee, et~al.]{lugaresi2019mediapipe}
Camillo Lugaresi, Jiuqiang Tang, Hadon Nash, Chris McClanahan, Esha Uboweja,
  Michael Hays, Fan Zhang, Chuo-Ling Chang, Ming~Guang Yong, Juhyun Lee, et~al.
\newblock Mediapipe: A framework for building perception pipelines.
\newblock \emph{arXiv preprint arXiv:1906.08172}, 2019.

\bibitem[Ma et~al.(2024)Ma, He, Cun, Wang, Chen, Li, and Chen]{ma2024follow}
Yue Ma, Yingqing He, Xiaodong Cun, Xintao Wang, Siran Chen, Xiu Li, and Qifeng
  Chen.
\newblock Follow your pose: Pose-guided text-to-video generation using
  pose-free videos.
\newblock In \emph{AAAI}, pages 4117--4125, 2024.

\bibitem[Meng et~al.(2021)Meng, He, Song, Song, Wu, Zhu, and Ermon]{sdedit}
Chenlin Meng, Yutong He, Yang Song, Jiaming Song, Jiajun Wu, Jun-Yan Zhu, and
  Stefano Ermon.
\newblock Sdedit: Guided image synthesis and editing with stochastic
  differential equations.
\newblock \emph{arXiv preprint arXiv:2108.01073}, 2021.

\bibitem[Mokady et~al.(2022)Mokady, Tov, Yarom, Lang, Mosseri, Dekel, Cohen-Or,
  and Irani]{mokady2022self}
Ron Mokady, Omer Tov, Michal Yarom, Oran Lang, Inbar Mosseri, Tali Dekel,
  Daniel Cohen-Or, and Michal Irani.
\newblock Self-distilled stylegan: Towards generation from internet photos.
\newblock In \emph{ACM SIGGRAPH}, pages 1--9, 2022.

\bibitem[Mokady et~al.(2023)Mokady, Hertz, Aberman, Pritch, and Cohen-Or]{null}
Ron Mokady, Amir Hertz, Kfir Aberman, Yael Pritch, and Daniel Cohen-Or.
\newblock Null-text inversion for editing real images using guided diffusion
  models.
\newblock In \emph{CVPR}, pages 6038--6047, 2023.

\bibitem[Nichol et~al.(2021)Nichol, Dhariwal, Ramesh, Shyam, Mishkin, McGrew,
  Sutskever, and Chen]{nichol2021glide}
Alex Nichol, Prafulla Dhariwal, Aditya Ramesh, Pranav Shyam, Pamela Mishkin,
  Bob McGrew, Ilya Sutskever, and Mark Chen.
\newblock Glide: Towards photorealistic image generation and editing with
  text-guided diffusion models.
\newblock \emph{arXiv preprint arXiv:2112.10741}, 2021.

\bibitem[Ouyang et~al.(2024)Ouyang, Wang, Xiao, Bai, Zhang, Zheng, Zhou, Chen,
  and Shen]{codef}
Hao Ouyang, Qiuyu Wang, Yuxi Xiao, Qingyan Bai, Juntao Zhang, Kecheng Zheng,
  Xiaowei Zhou, Qifeng Chen, and Yujun Shen.
\newblock Codef: Content deformation fields for temporally consistent video
  processing.
\newblock \emph{CVPR}, 2024.

\bibitem[Pang et~al.(2021)Pang, Lin, Qin, and Chen]{pang2021image}
Yingxue Pang, Jianxin Lin, Tao Qin, and Zhibo Chen.
\newblock Image-to-image translation: Methods and applications.
\newblock \emph{IEEE Transactions on Multimedia}, 24:\penalty0 3859--3881,
  2021.

\bibitem[Peng et~al.(2024)Peng, Wang, Zhang, Li, Yang, and
  Jia]{peng2024controlnext}
Bohao Peng, Jian Wang, Yuechen Zhang, Wenbo Li, Ming-Chang Yang, and Jiaya Jia.
\newblock Controlnext: Powerful and efficient control for image and video
  generation.
\newblock \emph{arXiv preprint arXiv:2408.06070}, 2024.

\bibitem[Qi et~al.(2023)Qi, Cun, Zhang, Lei, Wang, Shan, and Chen]{fatezero}
Chenyang Qi, Xiaodong Cun, Yong Zhang, Chenyang Lei, Xintao Wang, Ying Shan,
  and Qifeng Chen.
\newblock Fatezero: Fusing attentions for zero-shot text-based video editing.
\newblock \emph{arXiv preprint arXiv:2303.09535}, 2023.

\bibitem[Radford et~al.(2021)Radford, Kim, Hallacy, Ramesh, Goh, Agarwal,
  Sastry, Askell, Mishkin, Clark, et~al.]{radford2021learning}
Alec Radford, Jong~Wook Kim, Chris Hallacy, Aditya Ramesh, Gabriel Goh,
  Sandhini Agarwal, Girish Sastry, Amanda Askell, Pamela Mishkin, Jack Clark,
  et~al.
\newblock Learning transferable visual models from natural language
  supervision.
\newblock In \emph{International conference on machine learning}, pages
  8748--8763. PMLR, 2021.

\bibitem[Ramesh et~al.(2022)Ramesh, Dhariwal, Nichol, Chu, and Chen]{dalle2}
Aditya Ramesh, Prafulla Dhariwal, Alex Nichol, Casey Chu, and Mark Chen.
\newblock Hierarchical text-conditional image generation with clip latents.
\newblock \emph{arXiv preprint arXiv:2204.06125}, 2022.

\bibitem[Rombach et~al.(2022)Rombach, Blattmann, Lorenz, Esser, and Ommer]{ldm}
Robin Rombach, Andreas Blattmann, Dominik Lorenz, Patrick Esser, and Bj{\"o}rn
  Ommer.
\newblock High-resolution image synthesis with latent diffusion models.
\newblock In \emph{CVPR}, pages 10684--10695, 2022.

\bibitem[Ruiz et~al.(2023)Ruiz, Li, Jampani, Pritch, Rubinstein, and
  Aberman]{dreambooth}
Nataniel Ruiz, Yuanzhen Li, Varun Jampani, Yael Pritch, Michael Rubinstein, and
  Kfir Aberman.
\newblock Dreambooth: Fine tuning text-to-image diffusion models for
  subject-driven generation.
\newblock In \emph{CVPR}, pages 22500--22510, 2023.

\bibitem[Saharia et~al.(2022)Saharia, Chan, Saxena, Li, Whang, Denton,
  Ghasemipour, Gontijo~Lopes, Karagol~Ayan, Salimans, et~al.]{imagen}
Chitwan Saharia, William Chan, Saurabh Saxena, Lala Li, Jay Whang, Emily~L
  Denton, Kamyar Ghasemipour, Raphael Gontijo~Lopes, Burcu Karagol~Ayan, Tim
  Salimans, et~al.
\newblock Photorealistic text-to-image diffusion models with deep language
  understanding.
\newblock \emph{NeurIPS}, 35:\penalty0 36479--36494, 2022.

\bibitem[Singer et~al.(2022)Singer, Polyak, Hayes, Yin, An, Zhang, Hu, Yang,
  Ashual, Gafni, et~al.]{singer2022make}
Uriel Singer, Adam Polyak, Thomas Hayes, Xi Yin, Jie An, Songyang Zhang, Qiyuan
  Hu, Harry Yang, Oron Ashual, Oran Gafni, et~al.
\newblock Make-a-video: Text-to-video generation without text-video data.
\newblock \emph{arXiv preprint arXiv:2209.14792}, 2022.

\bibitem[Song et~al.(2020)Song, Meng, and Ermon]{ddim}
Jiaming Song, Chenlin Meng, and Stefano Ermon.
\newblock Denoising diffusion implicit models.
\newblock \emph{arXiv preprint arXiv:2010.02502}, 2020.

\bibitem[Teed and Deng(2020)]{raft}
Zachary Teed and Jia Deng.
\newblock Raft: Recurrent all-pairs field transforms for optical flow.
\newblock In \emph{ECCV}, pages 402--419, 2020.

\bibitem[Tumanyan et~al.(2023)Tumanyan, Geyer, Bagon, and Dekel]{pnp}
Narek Tumanyan, Michal Geyer, Shai Bagon, and Tali Dekel.
\newblock Plug-and-play diffusion features for text-driven image-to-image
  translation.
\newblock In \emph{CVPR}, pages 1921--1930, 2023.

\bibitem[Wang et~al.(2023)Wang, Yuan, Chen, Zhang, Wang, and
  Zhang]{wang2023modelscope}
Jiuniu Wang, Hangjie Yuan, Dayou Chen, Yingya Zhang, Xiang Wang, and Shiwei
  Zhang.
\newblock Modelscope text-to-video technical report.
\newblock \emph{arXiv preprint arXiv:2308.06571}, 2023.

\bibitem[Wang et~al.(2024)Wang, Park, Zhou, Shechtman, and Zhang]{jumpcut}
Xiaojuan Wang, Taesung Park, Yang Zhou, Eli Shechtman, and Richard Zhang.
\newblock Jump cut smoothing for talking heads.
\newblock \emph{arXiv preprint arXiv:2401.04718}, 2024.

\bibitem[Wang et~al.(2004)Wang, Bovik, Sheikh, and Simoncelli]{SSIM}
Zhou Wang, Alan~C Bovik, Hamid~R Sheikh, and Eero~P Simoncelli.
\newblock Image quality assessment: From error visibility to structural
  similarity.
\newblock \emph{IEEE Transactions on Image Processing}, 13\penalty0
  (4):\penalty0 600--612, 2004.

\bibitem[Wu et~al.(2023)Wu, Ge, Wang, Lei, Gu, Shi, Hsu, Shan, Qie, and
  Shou]{tune-a-video}
Jay~Zhangjie Wu, Yixiao Ge, Xintao Wang, Stan~Weixian Lei, Yuchao Gu, Yufei
  Shi, Wynne Hsu, Ying Shan, Xiaohu Qie, and Mike~Zheng Shou.
\newblock Tune-a-video: One-shot tuning of image diffusion models for
  text-to-video generation.
\newblock In \emph{ICCV}, pages 7623--7633, 2023.

\bibitem[Wu et~al.(2020)Wu, Zhou, Wilson, Xing, and Hu]{wu2020improving}
Yue Wu, Pan Zhou, Andrew~G Wilson, Eric Xing, and Zhiting Hu.
\newblock Improving gan training with probability ratio clipping and sample
  reweighting.
\newblock \emph{NeurIPS}, 33:\penalty0 5729--5740, 2020.

\bibitem[Xiao et~al.(2023)Xiao, Yin, Freeman, Durand, and Han]{fastcomposer}
Guangxuan Xiao, Tianwei Yin, William~T Freeman, Fr{\'e}do Durand, and Song Han.
\newblock Fastcomposer: Tuning-free multi-subject image generation with
  localized attention.
\newblock \emph{arXiv preprint arXiv:2305.10431}, 2023.

\bibitem[Xu et~al.(2024)Xu, Zhang, Liew, Yan, Liu, Zhang, Feng, and
  Shou]{xu2024magicanimate}
Zhongcong Xu, Jianfeng Zhang, Jun~Hao Liew, Hanshu Yan, Jia-Wei Liu, Chenxu
  Zhang, Jiashi Feng, and Mike~Zheng Shou.
\newblock Magicanimate: Temporally consistent human image animation using
  diffusion model.
\newblock In \emph{CVPR}, pages 1481--1490, 2024.

\bibitem[Yang et~al.(2023{\natexlab{a}})Yang, Zhou, Liu, and
  Loy]{yang2023rerender}
Shuai Yang, Yifan Zhou, Ziwei Liu, and Chen~Change Loy.
\newblock Rerender a video: Zero-shot text-guided video-to-video translation.
\newblock In \emph{SIGGRAPH Asia 2023 Conference Papers}, pages 1--11,
  2023{\natexlab{a}}.

\bibitem[Yang et~al.(2023{\natexlab{b}})Yang, Zeng, Yuan, and
  Li]{yang2023effective}
Zhendong Yang, Ailing Zeng, Chun Yuan, and Yu Li.
\newblock Effective whole-body pose estimation with two-stages distillation.
\newblock In \emph{ICCV}, pages 4210--4220, 2023{\natexlab{b}}.

\bibitem[Yatim et~al.(2023)Yatim, Fridman, Tal, Kasten, and
  Dekel]{yatim2023space}
Danah Yatim, Rafail Fridman, Omer~Bar Tal, Yoni Kasten, and Tali Dekel.
\newblock Space-time diffusion features for zero-shot text-driven motion
  transfer.
\newblock \emph{arXiv preprint arXiv:2311.17009}, 2023.

\bibitem[Yu et~al.(2023)Yu, Po, Cheung, Zhao, Xue, and Li]{bdmm}
Wing-Yin Yu, Lai-Man Po, Ray~CC Cheung, Yuzhi Zhao, Yu Xue, and Kun Li.
\newblock Bidirectionally deformable motion modulation for video-based human
  pose transfer.
\newblock In \emph{ICCV}, pages 7502--7512, 2023.

\bibitem[Yuan et~al.(2023)Yuan, Cun, Zhang, Li, Qi, Wang, Shan, and
  Zheng]{celeb}
Ge Yuan, Xiaodong Cun, Yong Zhang, Maomao Li, Chenyang Qi, Xintao Wang, Ying
  Shan, and Huicheng Zheng.
\newblock Inserting anybody in diffusion models via celeb basis.
\newblock \emph{NeurIPS}, 2023.

\bibitem[Zablotskaia et~al.(2019)Zablotskaia, Siarohin, Zhao, and
  Sigal]{zablotskaia2019dwnet}
Polina Zablotskaia, Aliaksandr Siarohin, Bo Zhao, and Leonid Sigal.
\newblock Dwnet: Dense warp-based network for pose-guided human video
  generation.
\newblock \emph{arXiv preprint arXiv:1910.09139}, 2019.

\bibitem[Zhang et~al.(2023{\natexlab{a}})Zhang, Mo, Chen, Sun, and
  Su]{Zhang2023MagicBrush}
Kai Zhang, Lingbo Mo, Wenhu Chen, Huan Sun, and Yu Su.
\newblock Magicbrush: A manually annotated dataset for instruction-guided image
  editing.
\newblock In \emph{NeurIPS}, 2023{\natexlab{a}}.

\bibitem[Zhang et~al.(2023{\natexlab{b}})Zhang, Rao, and Agrawala]{controlnet}
Lvmin Zhang, Anyi Rao, and Maneesh Agrawala.
\newblock Adding conditional control to text-to-image diffusion models.
\newblock In \emph{ICCV}, pages 3836--3847, 2023{\natexlab{b}}.

\bibitem[Zhang et~al.(2018)Zhang, Isola, Efros, Shechtman, and Wang]{lpips}
Richard Zhang, Phillip Isola, Alexei~A Efros, Eli Shechtman, and Oliver Wang.
\newblock The unreasonable effectiveness of deep features as a perceptual
  metric.
\newblock In \emph{CVPR}, pages 586--595, 2018.

\bibitem[Zhang et~al.(2023{\natexlab{c}})Zhang, Wang, Zhang, Zhao, Yuan, Qin,
  Wang, Zhao, and Zhou]{zhang2023i2vgen}
Shiwei Zhang, Jiayu Wang, Yingya Zhang, Kang Zhao, Hangjie Yuan, Zhiwu Qin,
  Xiang Wang, Deli Zhao, and Jingren Zhou.
\newblock I2vgen-xl: High-quality image-to-video synthesis via cascaded
  diffusion models.
\newblock \emph{arXiv preprint arXiv:2311.04145}, 2023{\natexlab{c}}.

\bibitem[Zhang et~al.(2021)Zhang, Li, Ding, and Fan]{zhang2021HDTF}
Zhimeng Zhang, Lincheng Li, Yu Ding, and Changjie Fan.
\newblock Flow-guided one-shot talking face generation with a high-resolution
  audio-visual dataset.
\newblock In \emph{CVPR}, pages 3661--3670, 2021.

\bibitem[Zhou et~al.(2022)Zhou, Wang, Yan, Lv, Zhu, and
  Feng]{zhou2022magicvideo}
Daquan Zhou, Weimin Wang, Hanshu Yan, Weiwei Lv, Yizhe Zhu, and Jiashi Feng.
\newblock Magicvideo: Efficient video generation with latent diffusion models.
\newblock \emph{arXiv preprint arXiv:2211.11018}, 2022.

\bibitem[Zhu et~al.(2022)Zhu, Wu, Zhu, Jiang, Tang, Zhang, Liu, and
  Loy]{zhu2022celebvhq}
Hao Zhu, Wayne Wu, Wentao Zhu, Liming Jiang, Siwei Tang, Li Zhang, Ziwei Liu,
  and Chen~Change Loy.
\newblock {CelebV-HQ}: A large-scale video facial attributes dataset.
\newblock In \emph{ECCV}, 2022.

\end{thebibliography}
